\documentclass[10pt,twocolumn,letterpaper]{article}

\usepackage{wacv}              

\usepackage{adjustbox}

\definecolor{wacvblue}{rgb}{0.21,0.49,0.74}
\usepackage[pagebackref,breaklinks,colorlinks,allcolors=wacvblue]{hyperref}

\def\wacvPaperID{*****} 
\def\confName{WACV}
\def\confYear{2027}

\title{TriINR: Unsupervised Atmospheric Turbulence Mitigation with\\ Triple Implicit Neural Representations}

\author{ Yuka Ogino, Rikuto Otsuka, Yuho Shoji, Takuya Ogawa, Takahiro Toizumi, Atsushi Ito\\ NEC Corporation\\Japan\\ 
{\tt\small \{yogino, otsuka-rikuto, yuho-shoji, takuya\_ogawa, t-toizumi\_ct, ito-atsushi\}@nec.com}
}

\begin{document}
\maketitle
\begin{abstract}
Atmospheric turbulence degrades image quality in long-range imaging. The degradation is generally modeled with two spatio-temporally varying components, tilt and blur. 
Recent unsupervised turbulence mitigation methods have demonstrated the effectiveness of representing a sharp image as an implicit neural representation (INR). 
However, their tilt and blur representations remain frame-dependent, and the number of optimization parameters grows with the number of input frames. 
To address this limitation, we propose TriINR, an unsupervised turbulence mitigation method that recovers a sharp image from multiple degraded frames. TriINR represents the sharp image, tilt, and blur as three implicit neural representations. By modeling tilt and blur as spatio-temporal INRs, TriINR achieves frame-count-independent degradation representation and memory-efficient optimization. 
We further incorporate physical and statistical properties of atmospheric turbulence as constraints on the INRs to improve restoration accuracy. 
Experiments on synthesized turbulence data with three degradation strengths show that TriINR restores images more robustly across varying turbulence strengths than prior TM methods and achieves the highest accuracy among the evaluated unsupervised methods. Our method also demonstrates robustness on real long-range scenes containing additional non-ideal motion.
\end{abstract}

\section{Introduction}
\label{sec:intro}
Atmospheric turbulence degrades image quality in long-range imaging. Temperature gradients and wind cause fluctuations of the refractive index, which distort the phase of light propagating through the atmosphere. The degradation affects not only image quality but also downstream tasks such as recognition and detection in surveillance applications. The phase distortion decomposes into two dominant components: a pixel-wise geometric shift (tilt) and a spatially varying blur~\cite{chan2022}. Both factors vary in space and time while preserving physical spatial correlations~\cite{noll1976, chan2022}. To alleviate this degradation, a variety of turbulence mitigation (TM) methods have been proposed~\cite{clear,tsrwgan,pirn,atddpm,tmt,datum,mambatm,nert,nertarxiv,tmfs}.

Based on their optimization paradigm, TM methods are generally categorized into supervised and unsupervised approaches. Supervised methods~\cite{tsrwgan,pirn,atddpm,tmt,datum,mambatm} learn an image transformation function from degraded frames to a sharp image using prepared data consisting of degraded frames paired with sharp images. The prepared data are typically generated by turbulence simulators~\cite{chimitt2020, mao2021, hardie2017} due to the difficulty of capturing pixel-aligned pairs of turbulence-degraded and sharp images in the real world. These methods work well within the training distribution, but they inherit biases from the simulator parameters and the base image distribution. Under out-of-distribution turbulence strengths or image domains, they hallucinate textures~\cite{tmfs}. Unsupervised methods, in contrast, reconstruct a sharp image only from degraded frames, based on either statistical properties or an explicit tilt-plus-blur degradation model. They do not rely on pretrained restoration priors and therefore avoid train–test shifts.

\begin{figure}
    \centering
    \includegraphics[width=0.95\linewidth]{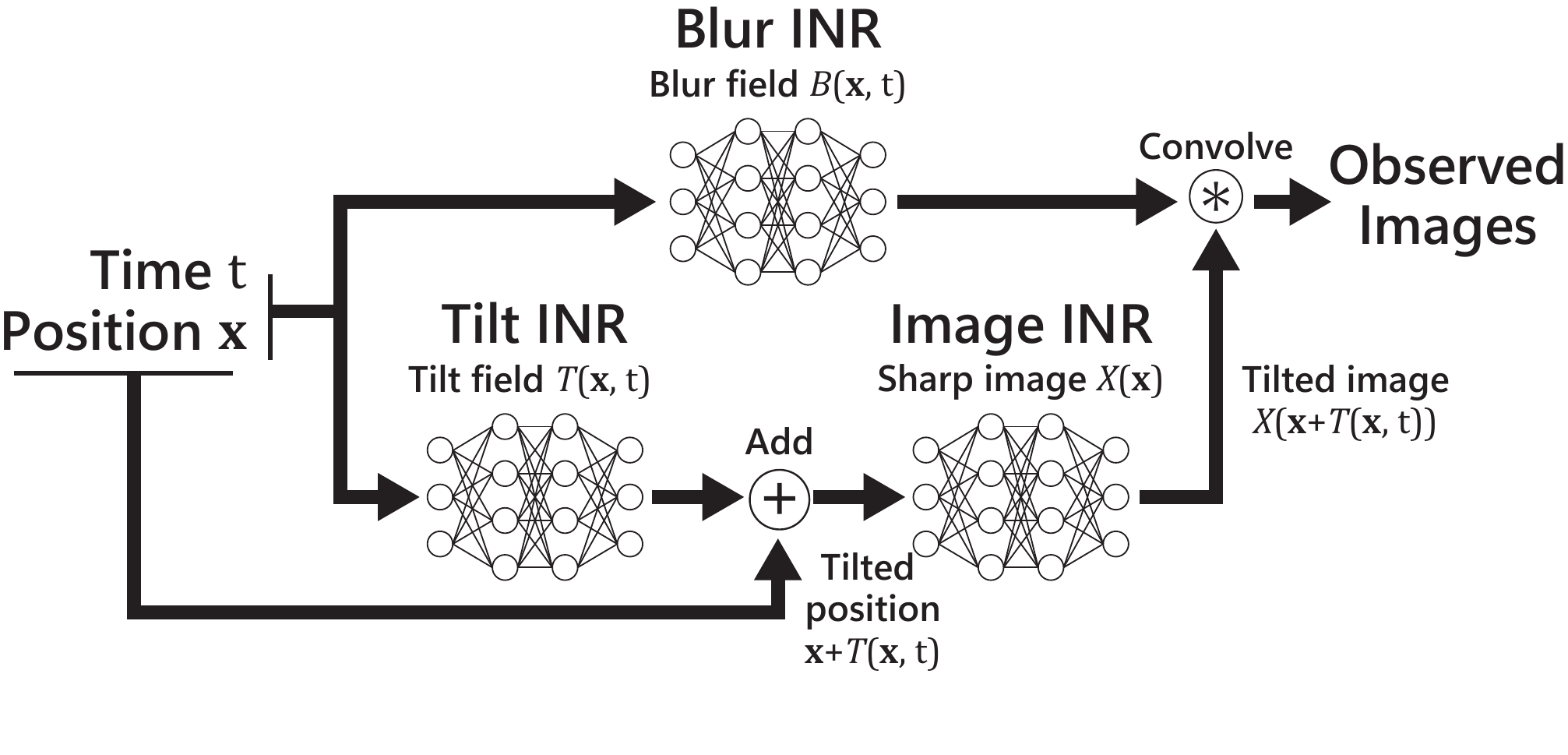}
    \caption{Overview of TriINR. TriINR reformulates the sharp image and the degradations (tilt and blur) as three continuous implicit neural representations and jointly optimizes them from multiple turbulence-degraded observations. The tilt and the blur are each modeled as a single spatio-temporal neural field shared across all frames, yielding a frame-count-independent representation.}
    \label{fig:teaser}
\end{figure}

Among unsupervised methods, statistics-based methods exploit the property that phase fluctuations of atmospheric turbulence behave as a random process. Noll showed that the phase fluctuations caused by turbulence can be represented by Zernike coefficients modeled as Gaussian random variables with Kolmogorov variance~\cite{noll1976}. Leveraging this property, Lucky imaging~\cite{fried1978, aubailly2009} exploits the fact that, over many frames, a "lucky frame" with a small instantaneous phase distortion arises, and selects sharp regions of such frames and applies weighted averaging to them. The restoration quality of Lucky imaging is bounded by the narrowest PSF present in the input frames~\cite{chanchimitt2024}.
 
In contrast, other unsupervised methods model the image degradation as a composition of tilt and blur, and optimize a sharp image that fits the tilt-and-blur model against multiple observations. This approach recovers a sharp image closer to the diffraction limit than statistical selection alone can achieve~\cite{zhumilanfar2013}. This formulation extends multi-frame blind deconvolution (MFBD), which jointly optimizes a sharp image, blur kernels, and tilt fields~\cite{chan2022}. The optimization is inherently ill-posed and requires design of the optimization framework and regularization. Prior work introduces image priors as regularization, or imposes physical constraints on the degradation to reduce the number of free parameters.

Recently, some unsupervised methods represent a sharp image as an implicit neural representation (INR) and achieve high-fidelity per-scene restoration~\cite{ndir, nert, tmfs}. Replacing the tensor-based image representation with an INR compresses the parameters of the image representation and provides an implicit regularization effect. However, existing INR-based methods represent only the sharp image as an INR, while tilt and blur are still parameterized as per-frame tensors or per-frame CNNs. The number of optimization parameters therefore grows linearly with the number of frames. We argue that tilt and blur are also continuous fields over space and time, and therefore can be represented compactly using spatio-temporal implicit neural representations. This representation shares parameters across all frames and enables coordinate sampling during optimization, reducing both parameter growth and memory consumption.

In this paper, we address the frame-dependent degradation representations used in existing INR-based TM methods. We propose TriINR, which reformulates the sharp image, tilt field, and blur field as continuous neural fields and jointly optimizes them from observed images. 

The contributions of this paper are three-fold:

\begin{itemize}
    \item {\bf Frame-count-independent degradation representation using triple INRs}: We represent the sharp image, tilt field, and blur field as continuous implicit neural representations. Our formulation represents atmospheric degradation as shared spatio-temporal neural fields for frame-count-independent representation.
    \item {\bf Scalable optimization for high-resolution images}: Our formulation enables the optimizer to stochastically sample observation points at each step, keeping the memory cost independent of the image resolution.
    \item {\bf Physically and statistically grounded architecture of INRs}: We incorporate physical and statistical properties of atmospheric turbulence as constraints in the architecture of the three INRs. In particular, we enforce a zero-temporal-mean constraint on the tilt INR field.
\end{itemize}
Experiments across three simulated turbulence strengths show that TriINR is robust to varying turbulence conditions and achieves the highest restoration accuracy among the evaluated unsupervised methods.  Qualitative results further demonstrate robustness on real long-range scenes containing additional non-ideal motion.

\section{Related work}
\label{sec:related}
\begin{table*}[t]
\centering
\small
\caption{Parameter and representation comparison of unsupervised INR-based turbulence mitigation methods. Unlike prior methods, whose parameter count grows with the frame count and image size, TriINR keeps a constant parameter count. PE denotes positional encoding, and GF denotes Gaussian Fourier features~\cite{fourier_features}.}
\label{tab:param_compare}
\setlength{\tabcolsep}{4pt}
\renewcommand{\arraystretch}{1.15}
\setlength{\tabcolsep}{6pt}
\renewcommand{\arraystretch}{1.15}
\begin{tabular}{@{}l c c c@{}}
\toprule
& NeRT~\cite{nert}& TMFS~\cite{tmfs} & \textbf{TriINR (ours)}  \\
\midrule
Tilt represented as INR & \checkmark & $\times$ & \checkmark  \\
Blur represented as INR  & $\times$ & \checkmark & \checkmark   \\
Image represented as INR  & \checkmark  & \checkmark  & \checkmark \\
\midrule
Parameter count scaling & $O(N)$ & $O(H W N)$ & $\boldsymbol{O(1)}$ \\
\footnotesize \#Params ($H{=}W{=}256$, $N{=}20$) & $\sim 3.8$M & $\sim 200$M & $\sim \mathbf{0.4M}$ \\
\midrule
\footnotesize  Tilt details
    &\footnotesize  Frame-dependent INRs
    &\footnotesize  Frame-dependent tensors
    &\footnotesize  \textbf{Frame-independent INR} \\
    &\footnotesize  Grid deformer~\cite{ndir}, one CNN / frame
    &\footnotesize  Noise map + correlation
    &\footnotesize  PE (GF) + MLP, $(x,y,t)$ input \\
\midrule
\footnotesize Blur details
    &\footnotesize  Frame-dependent tensors
    &\footnotesize  Frame- \& resolution-dependent tensors
    &\footnotesize  \textbf{Frame-independent INR} \\
    &\footnotesize  Zernike vector map
    &\footnotesize  Noise map + correlation parameters
    &\footnotesize  PE (GF) + MLP, $(x,y,t)$ input \\
\midrule
\footnotesize Image details
    &\footnotesize  Single INR
    &\footnotesize  Single INR
    &\footnotesize  Single INR \\
    &\footnotesize  PE (GF) + MLP
    &\footnotesize  MLP (PE unspecified)
    &\footnotesize  PE (Hashgrid~\cite{instantngp}) + MLP \\
\bottomrule
\end{tabular}
\end{table*}

\subsection{Atmospheric turbulence model}
Atmospheric turbulence is modeled as fluctuations of the optical wavefront phase. It appears on the image as a pixel-wise varying convolution. Recent computer vision research has proposed simulators of atmospheric turbulence for the study and evaluation of image recognition and restoration methods. Most approaches represent the turbulence-induced phase distortion using Zernike coefficients, and describe the turbulence via the Noll covariance matrix~\cite{noll1976} derived from the Kolmogorov turbulence power spectrum. When the Zernike-represented phase distortion is transformed to the spatial domain, its components separate into a tilt component that corresponds to a pixel-wise geometric shift, and a blur component that represents the pixel-wise blur as a point spread function (PSF). Because transforming per-pixel Zernike coefficients to the spatial domain is computationally expensive, Mao et al.~\cite{mao2021} proposed to implement the Phase-to-Space (P2S) transform for the blur component as a small MLP.

\subsection{Turbulence mitigation for degraded images}
Turbulence Mitigation (TM) methods have been proposed based on these degradation models. TM methods are broadly categorized into supervised and unsupervised approaches. Supervised methods pretrain a mitigation function on paired clean and degraded images obtained from simulation. Prior works have proposed diverse architectures, including GAN~\cite{tsrwgan},diffusion-model-based methods~\cite{atddpm,pirn}, transformer-based methods~\cite{tmt, datum}, and selective state space models~\cite{mambatm}. 

Unsupervised methods fit an image itself to the characteristics of the atmospheric turbulence model through weighted averaging or optimization. Unsupervised methods only require degraded images as input, and unlike supervised methods, they do not require any paired training data. Some methods use the statistical properties of the turbulence to select or weight-average across frames. Representative examples include lucky imaging~\cite{fried1978, aubailly2009} and CLEAR~\cite{clear}.
The restoration quality of these methods is bounded by the narrowest PSF present in the input frames. Other unsupervised methods explicitly models tilt and blur~\cite{shimizu2008,zhumilanfar2013}.  
They define a degradation model built from tilt and blur, and jointly (or in stages) optimize the sharp image, the tilt field, and the blur kernels. This formulation extends multi-frame blind deconvolution (MFBD). To mitigate the ill-posedness, prior methods have introduced image priors into the objective, or imposed physical constraints on the degradation to reduce the number of free parameters.

\begin{figure*}[t]
\centering
\setlength{\tabcolsep}{1pt}
\renewcommand{\arraystretch}{0}
\begin{tabular}{@{}ccc@{}}
\includegraphics[height=0.32\linewidth]{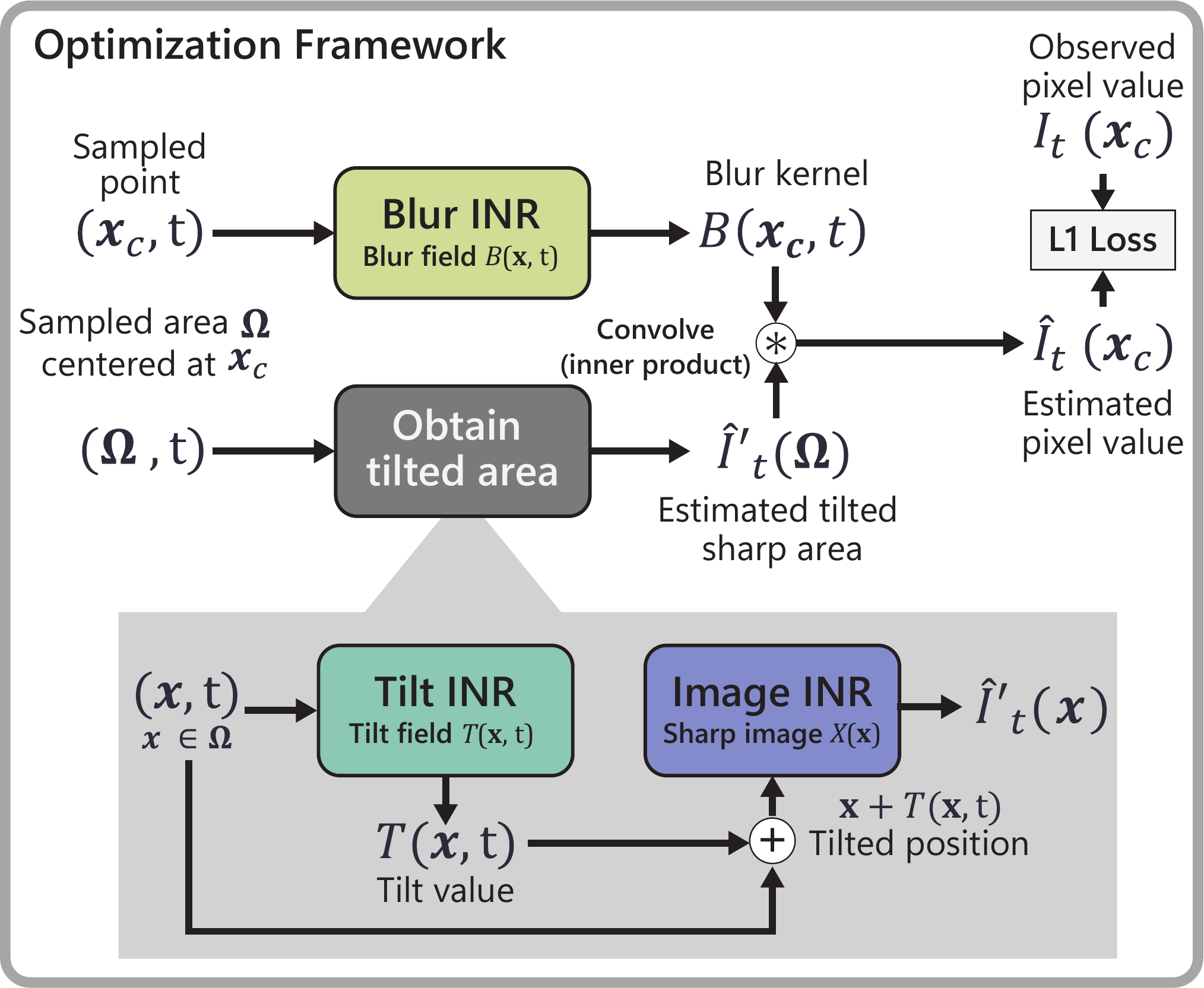}&\ &
\includegraphics[height=0.32\linewidth]{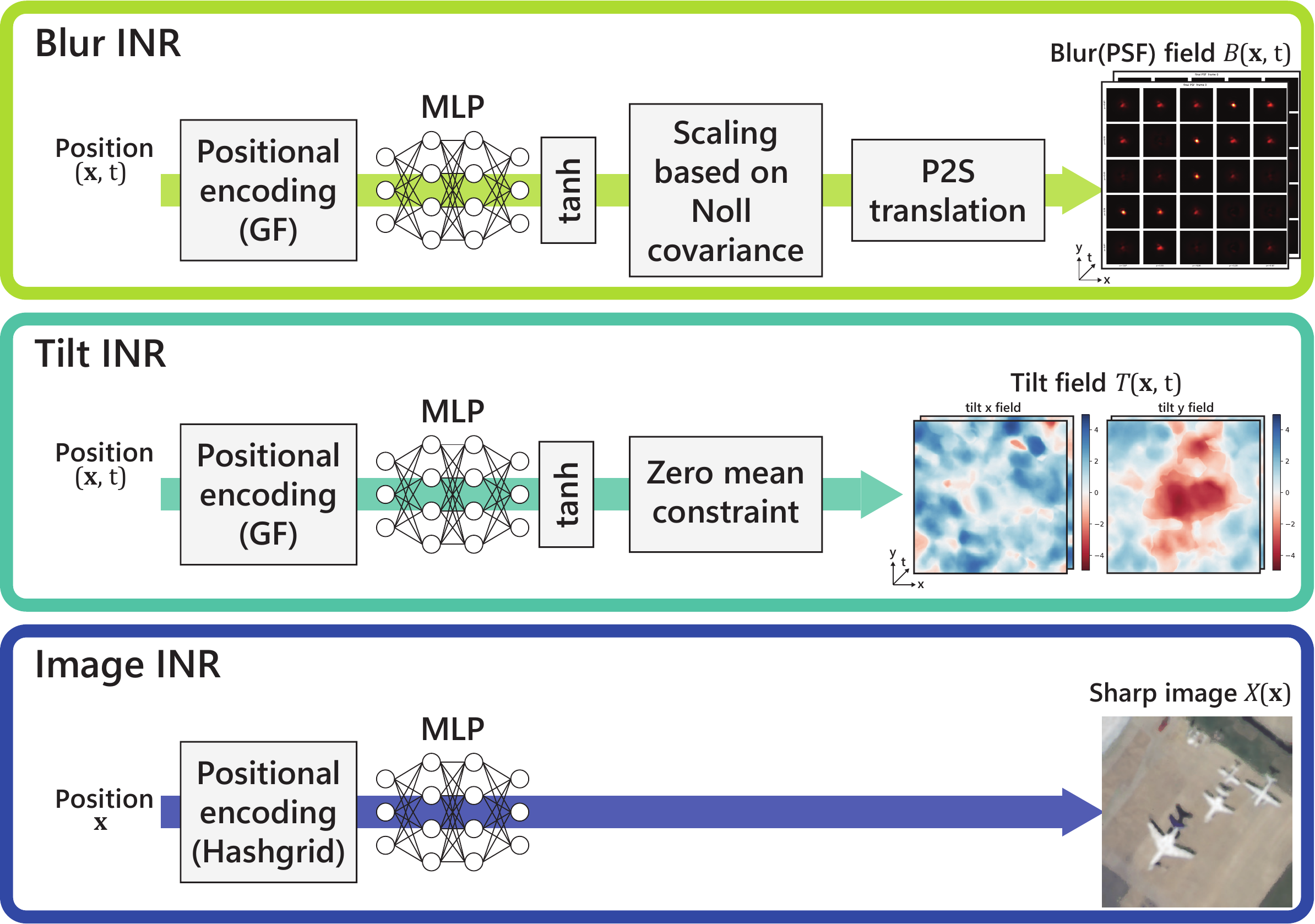}
\\[1pt]
\small (a) && (b)
\end{tabular}
\caption{Overview of the TriINR optimization framework. (a) Loss evaluation process in the second optimization stage. The triple INRs are jointly optimized through the atmospheric turbulence forward model. To improve scalability, only sampled observation coordinates and their local blur supports are evaluated at each optimization step. In the first stage, we optimize only the components shown in gray area. (b) Architecture of the three INRs. The Tilt INR and the Blur INR model the atmospheric degradation as spatio-temporal continuous fields shared across all input frames.}
\label{fig:arch}
\end{figure*}


\subsection{Unsupervised Turbulence Mitigation using INR}
Inspired by NeRF~\cite{nerf,siren,nerfies}, recent methods represent parts of the image degradation model as implicit neural representations (INRs) \cite{ndir,nert,nertarxiv,tmfs}. The NeRF architecture maps input coordinates to output colors and optimizes a 3D representation from multiple views, which is structurally analogous to restoring a sharp image from multiple degraded observations. By adopting techniques studied in the NeRF literature, such as high-frequency positional encoding~\cite{fourier_features,instantngp}, these methods can represent images at high resolution with few parameters.

NDIR~\cite{ndir} proposes a registration-based framework for non-rigid distortion removal in air and water. It represents the tilt as a CNN grid deformer, one per input frame, and the sharp image as a neural image generator. The grid deformer maps input coordinates to displaced coordinates, and requires an explicit initialization step that starts it close to the identity map. NeRT~\cite{nert} extends NDIR to atmospheric turbulence by adding a blur representation with the two INRs, using per-frame Zernike coefficient maps. TMFS~\cite{tmfs} instead emphasizes the physical fidelity of the atmospheric turbulence model. It keeps only the sharp image as a INR, and expresses the tilt and the blur as per-frame noise maps together with a frame-shared autocorrelation function derived from the turbulence model.

These prior works demonstrate the benefit of representing the sharp image as an INR, but leave room for improvement in the representation of tilt and blur. In particular, their tilt and blur representations scale with the number of input frames and with the image resolution. As a result, memory consumption increases for long sequences or high-resolution inputs. Table~\ref{tab:param_compare} summarizes the architecture and parameter count of INR-based turbulence mitigation methods. For the tilt, the grid deformer used by NeRT must be instantiated per frame, and TMFS attaches per-frame noise maps and correlation parameters to every input. For the blur, both NeRT and TMFS keep per-frame tensor representations; TMFS in particular also scales its parameter count with the image resolution, and therefore splits the input into $256 \times 256$-sized tiles.

In addition to this parameter-count issue, existing INR-based TM methods leave a scalability problem in the forward evaluation used during optimization. The public implementation of NeRT~\cite{nert,nertarxiv} evaluates the full-resolution spatial grid of every input frame at each optimization step, so the peak memory usage per step scales with the image resolution. In contrast, coordinate-based reconstruction methods such as NeRF~\cite{nerf} optimize large continuous fields by stochastically sampling a small subset of coordinates per step rather than evaluating the entire domain each iteration. We adopt this idea for the turbulence forward model: at each step we evaluate the model only at a sampled set of observation points and at the local support required for their blur convolution, enabling memory-efficient optimization on high-resolution inputs.

\section{Proposed method}
\label{sec:proposed}

We propose TriINR, which represents the sharp image, the tilt, and the blur as three independent implicit neural representations (INRs) (Fig.~\ref{fig:arch}). We incorporate the physical and statistical properties of atmospheric turbulence as constraints on the INRs. Each INR takes a spatial or spatio-temporal coordinate as input and outputs one of RGB values, a 2D tilt displacement, or Zernike coefficients that represent the PSF. As shown in Table~\ref{tab:param_compare}, our design reduces the parameter count compared with prior INR-based methods and does not scale with the frame number.

\subsection{Model and objective function}
\label{sec:model_obj}
Given a set of degraded image observations $\{I_t(\mathbf{x})\}_{t=1}^N$, where $\mathbf{x} = (x, y)$ denotes the spatial coordinate, our method jointly optimizes (i) a time-invariant sharp image $X_\theta(\mathbf{x})$, (ii) a spatially and temporally varying tilt field $\mathcal{T}_\phi(\mathbf{x}, t)$, and (iii) a spatially and temporally varying blur represented by per-pixel PSFs $\mathcal{B}_\omega(\mathbf{x}, t)$. We jointly optimize the parameters $\theta$, $\phi$, and $\omega$ of the three INRs to minimize the objective. Following the tilt-then-blur formulation~\cite{chan2022}, we express the modeled observation $\hat{I}_t(\mathbf{x})$ as a per-pixel PSF convolution applied to the tilted sharp image:
\begin{equation}
\hat{I}_t(\mathbf{x}) = \mathcal{B}_\omega(\mathbf{x}, t) \circledast X_\theta\bigl(\mathbf{x} + \mathcal{T}_\phi(\mathbf{x}, t)\bigr),
\end{equation}
where $\circledast$ denotes convolution over $\mathbf{x}$.

\begin{table*}[t]
\centering
\small
\caption{Quantitative comparison on four benchmarks. We report PSNR and SSIM
(higher is better) averaged over all scenes in each dataset. \textbf{Bold}: best among all methods.
\underline{Underlined}: best among unsupervised methods. NeRT on \emph{turb\_syn\_static} runs out of memory (OOM) at $512{\times}512$ resolution.}
\label{tab:quantitative}
\setlength{\tabcolsep}{4pt}
\renewcommand{\arraystretch}{1.1}
\begin{tabular}{@{}l |cc| c c c c| c c c c c c@{}}
\hline
&&& \multicolumn{4}{c|}{Supervised} & \multicolumn{6}{c}{Unsupervised} \\
&&& \multicolumn{2}{c}{TMT~\cite{tmt}} & \multicolumn{2}{c|}{DATUM~\cite{datum}} & \multicolumn{2}{c}{CLEAR~\cite{clear}} & \multicolumn{2}{c}{NeRT~\cite{nert}} & \multicolumn{2}{c}{\textbf{TriINR (ours)}} \\
Dataset &\#images&image size & PSNR & SSIM & PSNR & SSIM & PSNR & SSIM & PSNR & SSIM & PSNR & SSIM \\
\hline 
Weak turb.\ on LandUse&2100&256$\times$256 & 22.49 & 0.636 & 23.89 & 0.683 & 23.03 & 0.647 & 24.59 & 0.664 & \textbf{\underline{25.64}} & \textbf{\underline{0.707}} \\
Middle turb.\ on LandUse&2100&256$\times$256 & 26.23 & 0.783 & \textbf{27.34} & \textbf{0.817} & 24.03 & 0.711 & 24.06 & 0.643 & \underline{25.74} & \underline{0.725} \\
Strong turb.\ on LandUse&2100&256$\times$256 & 25.26 & 0.723 & \textbf{26.39} & \textbf{0.777} & 23.24 & 0.644 & 22.97 & 0.570 & \underline{24.41} & \underline{0.652} \\
\hline
\emph{turb\_syn\_static}~\cite{tmt}&1518&512$\times$512 & \textbf{27.97} & \textbf{0.849} & 27.52 & 0.849 & 24.76 & \underline{0.758} & OOM  & OOM& \underline{25.98} & {0.757} \\
\hline
\end{tabular}
\vspace{2pt}
\end{table*}

We define the objective as the $\ell_1$ reconstruction error over the observations following prior work~\cite{nert}. We further incorporate a statistical property of the atmospheric turbulence model as a zero-temporal-mean constraint on the tilt:
\begin{equation}
\min_{\theta, \phi, \omega} \sum_{\mathbf{x}, t} \bigl\| \hat{I}_t(\mathbf{x}) - I_t(\mathbf{x}) \bigr\|_1
\quad \text{s.t.} \quad \frac{1}{N}\sum_{t=1}^N \mathcal{T}_\phi(\mathbf{x}, t) = \mathbf{0}.
\end{equation}
We incorporate this zero-mean constraint into the architecture of the Tilt INR, as described later in Section~\ref{sec:tilt_inr}. 

\subsection{Optimization techniques for blur optimization}
\label{sec:opt_blur}

 We split the optimization into two stages for fast convergence following prior work~\cite{nert,tmfs}. In the first stage, only the Image INR $X_\theta$ and the Tilt INR $\mathcal{T}_\phi$ are optimized. The PSF is frozen at a delta function, and the objective simplifies to $\| X_\theta(\mathbf{x} + \mathcal{T}_\phi(\mathbf{x}, t)) - I_t(\mathbf{x}) \|_1$. In the second stage, all three INRs are optimized jointly using the full objective.

Following coordinate-based reconstruction methods such as NeRF~\cite{nerf}, we sample a subset of coordinates $(\mathbf{x}, t)$ from the observations at each step and update the INRs using the loss evaluated only at those samples. This keeps memory usage manageable at high image resolutions. In the first stage, we can freely sample $(\mathbf{x}, t)$ points and take an optimization step. In the second stage, however, evaluating the convolution at a sampled location $(\mathbf{x}_c, t)$ requires the tilted image over the sampled region of the blur kernel centered at $\mathbf{x}_c$. This is expressed as:
\begin{equation}
\hat{I}_t(\mathbf{x}_c) = \sum_{ \delta \in \Omega} \mathcal{B}_\omega(\mathbf{x}_c, t)[\delta] \cdot X_\theta\bigl(\mathbf{x}_c + \delta + \mathcal{T}_\phi(\mathbf{x}_c + \delta, t)\bigr),
\end{equation}
where $\Omega$ denotes the $d \times d$ neighborhood of $\mathbf{x}_c$. Although the Blur INR generates a $K \times K$ PSF, we only sample its central $d \times d$ region ($d < K$) to reduce the computational cost of the convolution evaluation.

Accordingly, as illustrated in Fig.~\ref{fig:arch} (a), we first sample $(\mathbf{x}_c, t)$ and obtain the full $K \times K$ PSF from $\mathcal{B}_\omega$. We then crop the central $d \times d$ region $\Omega$ and normalize its values to sum to one. Finally, we take the inner product of the normalized PSF with the tilted sharp image evaluated over $\Omega$. In practice, we sample fewer $(\mathbf{x}_c, t)$ points per iteration in the second stage than in the first stage and accumulate gradients across iterations.

\begin{figure*}[t]
\centering
\setlength{\tabcolsep}{1pt}
\renewcommand{\arraystretch}{0}
\begin{tabular}{@{}c@{\hspace{2pt}}ccccccc@{}}
\adjustbox{valign=c,rotate=90}{\small \quad \quad Weak} &
\includegraphics[width=0.13\linewidth]{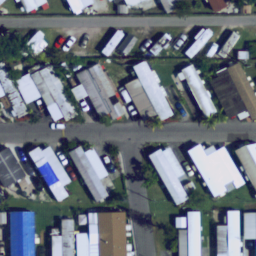}&
\includegraphics[width=0.13\linewidth]{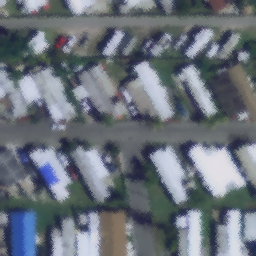}&
\includegraphics[width=0.13\linewidth]{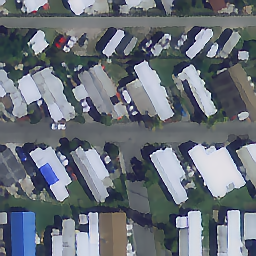}&
\includegraphics[width=0.13\linewidth]{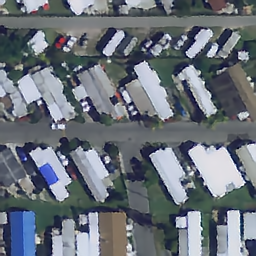}&
\includegraphics[width=0.13\linewidth]{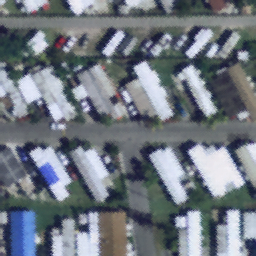}&
\includegraphics[width=0.13\linewidth]{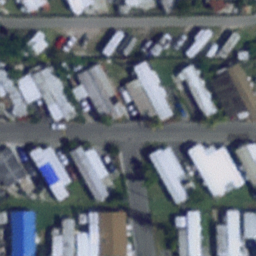}&
\includegraphics[width=0.13\linewidth]{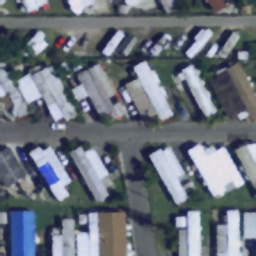}\\[1pt]
\adjustbox{valign=c,rotate=90}{\small \quad \quad Middle} &
\includegraphics[width=0.13\linewidth]{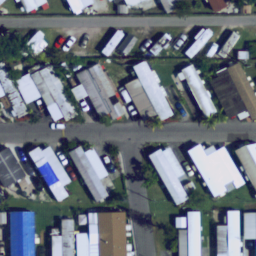}&
\includegraphics[width=0.13\linewidth]{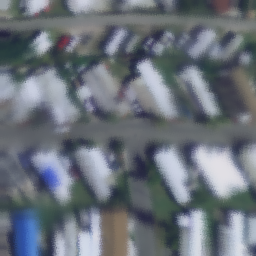}&
\includegraphics[width=0.13\linewidth]{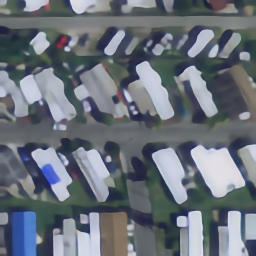}&
\includegraphics[width=0.13\linewidth]{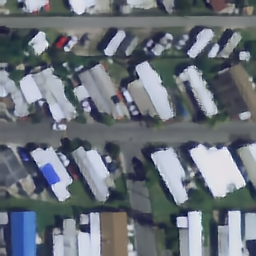}&
\includegraphics[width=0.13\linewidth]{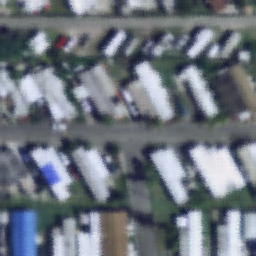}&
\includegraphics[width=0.13\linewidth]{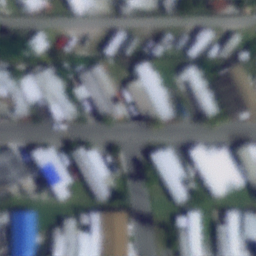}&
\includegraphics[width=0.13\linewidth]{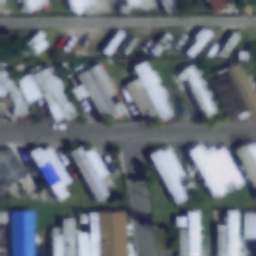}\\[1pt]
\adjustbox{valign=c,rotate=90}{\small \quad \quad Strong} &
\includegraphics[width=0.13\linewidth]{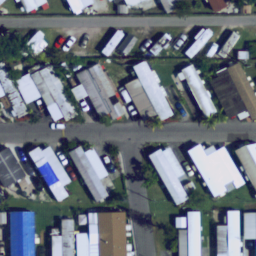}&
\includegraphics[width=0.13\linewidth]{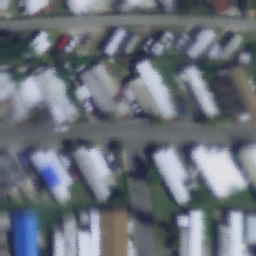}&
\includegraphics[width=0.13\linewidth]{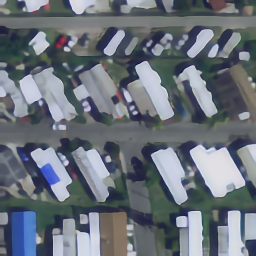}&
\includegraphics[width=0.13\linewidth]{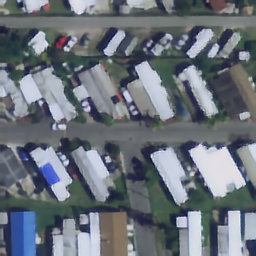}&
\includegraphics[width=0.13\linewidth]{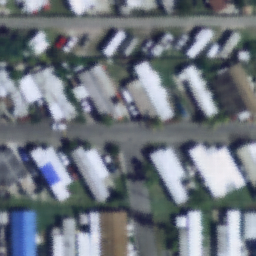}&
\includegraphics[width=0.13\linewidth]{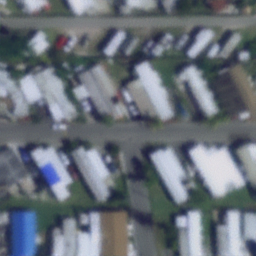}&
\includegraphics[width=0.13\linewidth]{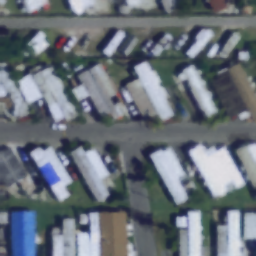}\\[1pt]
& \small GT & \small Input sample & \small TMT \cite{tmt} & \small DATUM \cite{datum} & \small CLEAR \cite{clear} & \small NeRT \cite{nert} & \small TriINR (ours) \\
\end{tabular}
\caption{Qualitative comparison on our LandUse-based synthetic dataset under three turbulence strengths (Weak / Middle / Strong; top to bottom), each restored from 20 input frames. Under Weak turbulence, supervised methods produce over-sharpening artifacts, whereas TriINR recovers the image faithfully. Across all three strengths, our method yields robust restoration.}
\label{fig:qual_landuse}
\end{figure*}

\subsection{Details of INR representations}

In our method, each INR consists of three components: a positional encoding (PE) of the input coordinate, an MLP, and constraints on the MLP output. The architecture of each INR is illustrated in Fig.~\ref{fig:arch} (b).

\subsubsection{Blur INR}
\label{sec:blur_zm}

The Blur INR $\mathcal{B}_\omega$ takes $(\mathbf{x}, t)$ as input and outputs a $K \times K$ PSF. It consists of two stages: an MLP that outputs a Zernike coefficient vector $\mathbf{z} \in \mathbb{R}^L$ ($L = 36$), and a pretrained Phase-to-Space (P2S) transform~\cite{mao2021} that converts the Zernike coefficients into a PSF. The MLP stage uses Gaussian Fourier positional encoding ($\mathrm{PE}_{GF}$) followed by an MLP. The P2S weights are frozen; TriINR itself does not require paired training of a restoration network.

Let $z(\mathbf{x}, t) \in \mathbb{R}^L$ denote the Zernike coefficient vector output by the MLP. We scale its output range by the diagonal $\sigma_{\mathrm{kol}} \in \mathbb{R}^L$ of the Noll covariance matrix~\cite{noll1976} derived from Kolmogorov turbulence statistics:
\begin{equation}
\label{eq:blur_scale}
z(\mathbf{x}, t) = \tanh\bigl(\mathrm{MLP}(\mathrm{PE}_{GF}(\mathbf{x}, t))\bigr) \cdot s_p \cdot \mathrm{diag}(\sigma_{\mathrm{kol}}),
\end{equation}
where $s_p$ is a coefficient scale.

The scaled Zernike coefficient vector is then passed to the pretrained MLP $\mathrm{P2S}(\cdot)$ and the Blur PSF is obtained as:
\begin{equation}
\mathcal{B}_\omega(\mathbf{x}, t) = \mathrm{P2S}\bigl(z(\mathbf{x}, t)\bigr).
\end{equation}
The PSF is generated at size $K \times K (K{=}33)$. As described in Section~\ref{sec:opt_blur}, when optimizing the blur, we crop the central $d \times d$ region and renormalize it to reduce computation.

We do not enforce a zero-temporal-mean constraint on the Blur INR's Zernike coefficients. This is because per-pixel PSF evaluation is computationally expensive, and the reduced sample size in the second optimization stage prevents the sample mean from reliably converging to the theoretical zero expectation. As shown in the ablation study (Section~\ref{sec:ablation}), omitting the blur zero-mean constraint yields better restoration accuracy.

\begin{table}[t]
\centering
\footnotesize
\caption{Network architecture of the three INRs in TriINR. 
$L$ denotes hashgrid levels, $F$ hashgrid feature channels per level, 
and $M$ Gaussian Fourier (GF) mapping size.}
\label{tab:arch}
\setlength{\tabcolsep}{5pt}
\begin{tabular}{lccc}
\toprule
& Base INR & Tilt INR & Blur INR \\
\midrule
Input        & $(x, y)$   & $(x, y, t)$  & $(x, y, t)$ \\
PE type             & Hashgrid & GF & GF \\
PE params       & $L{=}14$, $F{=}2$,& $M{=}64$ & $M{=}256$ \\
& hash table size $2^{14}$, &$\sigma{=}1.0$  &$\sigma{=} 4.0$ \\
& base resolution 16 &  & \\
MLP width           & 64              & 128           & 128 \\
MLP depth           & 4               & 4             & 4 \\
Activation          & ReLU            & ReLU          & ReLU \\
Output dim.         & 3         & 2 & 36 \\
\bottomrule
\end{tabular}
\end{table}

\subsubsection{Tilt INR}
\label{sec:tilt_inr}

The Tilt INR $\mathcal{T}_\phi$ takes a spatio-temporal coordinate $(x, y, t)$ as input and outputs a 2D tilt displacement $(u_x, u_y) \in \mathbb{R}^2$. It also uses Gaussian Fourier positional encoding ($\mathrm{PE}_{GF}$) followed by an MLP. The raw MLP output $\mathcal{T}'_\phi$ is scaled by tanh:
\begin{equation}
\label{eq:tilt_raw}
\mathcal{T}'_\phi(\mathbf{x}, t) = \tanh\bigl(\mathrm{MLP}(\mathrm{PE}_{GF}(\mathbf{x}, t))\bigr) \cdot s_t,
\end{equation}
where $s_t$ is a displacement scale.

We enforce zero temporal mean analytically within the forward pass by subtracting the temporal mean, yielding the final tilt output $\mathcal{T}_\phi$:
\begin{equation}
\mathcal{T}_\phi(\mathbf{x}, t) = \mathcal{T}'_\phi(\mathbf{x}, t) - \frac{1}{N}\sum_{t'=1}^N \mathcal{T}'_\phi(\mathbf{x}, t').
\end{equation}
This guarantees that the tilt representation satisfies zero temporal mean exactly at the architecture level.

\subsubsection{Image INR}
The Image INR $X_\theta$ takes a spatial coordinate $\mathbf{x} = (x, y)$ as input and outputs RGB values, representing the time-invariant sharp image as a continuous field. Unlike prior INR-based turbulence mitigation methods~\cite{nert} that use Gaussian Fourier positional encoding, we adopt hashgrid encoding~\cite{instantngp}. Hashgrid encoding provides learnable, multi-resolution features that enable faster convergence and higher-fidelity recovery of high-frequency structure.

\section{Experimental results}
\label{sec:experiment}

\label{sec:experiments}

In this section, we experimentally show that our method accurately recovers a sharp image from multiple turbulence-degraded frames. We first describe the implementation details and evaluation setup. We then present quantitative and qualitative results, and finally discuss the ablation study.

\subsection{Experimental setup}

\subsubsection{Implementation details}

All INRs are optimized with Adam ($\beta_1 = 0.9$, $\beta_2 = 0.99$) at a learning rate of $5 \times 10^{-3}$. The first stage runs for 1{,}000 iterations, sampling 16{,}384 $(\mathbf{x}, t)$ points per iteration without gradient accumulation. The second stage runs for 10{,}000 iterations, sampling 160 points per iteration and accumulating gradients over eight iterations, resulting in 1{,}280 points per gradient update. We set the displacement scale to $s_t = 0.2$ in Eq.~\ref{eq:tilt_raw} and the coefficient scale to $s_p = 5.0$ in Eq.~\ref{eq:blur_scale}. The size of $\Omega$ is $17 \times 17 (d = 17)$. Detailed architectures of each INR are provided in Table~\ref{tab:arch}. 

We use a specific initialization for the output layer of the Tilt INR and the Blur INR for stable optimization. The weights of the MLP's output layer are drawn from $\mathcal{U}(-\eta, +\eta)$ with $\eta_t = 10^{-3}$ (Tilt) and $\eta_p = 10^{-2}$ (Blur), and the corresponding biases are set to zero; all other layers, and the entire Image INR, use PyTorch's default (Kaiming uniform) initialization. These small values of $\eta$ keep the tanh activations following the MLPs in their linear regime at initialization; without this scaling, the activations would start in the saturated regime.
For the hashgrid encoding of the Image INR, we adopt a coarse-to-fine annealing schedule~\cite{barf}. During the first stage, only the coarser levels of the hashgrid are activated, and all levels are released at the beginning of the second stage. This prevents the Image INR from absorbing high-frequency components before the tilt and PSF estimates have stabilized.

\begin{figure*}[t]
\centering
\setlength{\tabcolsep}{1pt}
\renewcommand{\arraystretch}{0}
\begin{tabular}{@{}c@{\hspace{2pt}}cccccc@{}}
\adjustbox{valign=c,rotate=90}{\small \quad OTIS Pattern4} &
\includegraphics[width=0.13\linewidth]{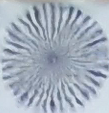}&
\includegraphics[width=0.13\linewidth]{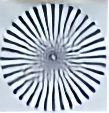}&
\includegraphics[width=0.13\linewidth]{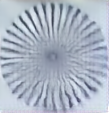}&
\includegraphics[width=0.13\linewidth]{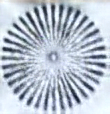}&
\includegraphics[width=0.13\linewidth]{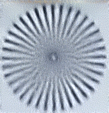}&
\includegraphics[width=0.13\linewidth]{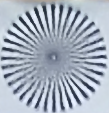}
\\[1pt]
\adjustbox{valign=c,rotate=90}{\small \quad OTIS Pattern14} &
\includegraphics[width=0.13\linewidth]{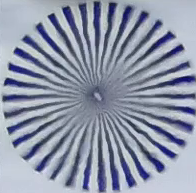}&
\includegraphics[width=0.13\linewidth]{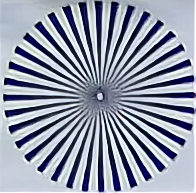}&
\includegraphics[width=0.13\linewidth]{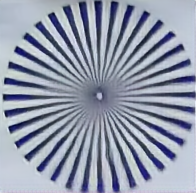}&
\includegraphics[width=0.13\linewidth]{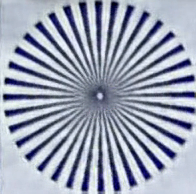}&
\includegraphics[width=0.13\linewidth]{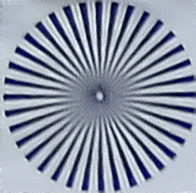}&
\includegraphics[width=0.13\linewidth]{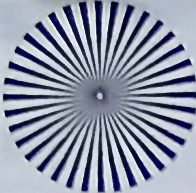}
\\[1pt]

& \small Input sample & \small TMT \cite{tmt} & \small DATUM \cite{datum} & \small CLEAR \cite{clear} & \small NeRT \cite{nert} & \small TriINR (ours) \\
\end{tabular}
\caption{Qualitative comparison on OTIS Pattern 4 and Pattern 14, each restored from 20 input frames. TriINR recovers the pattern with the highest fidelity among all compared methods.}
\label{fig:qual_otis}
\end{figure*}

\begin{figure}[t]
\centering
\setlength{\tabcolsep}{1pt}
\renewcommand{\arraystretch}{0}
\begin{tabular}{@{}ccc@{}}
\includegraphics[width=0.3\linewidth]{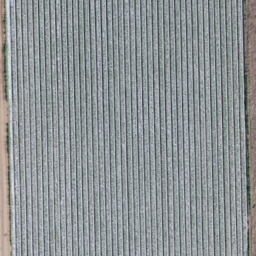}&
\includegraphics[width=0.3\linewidth]{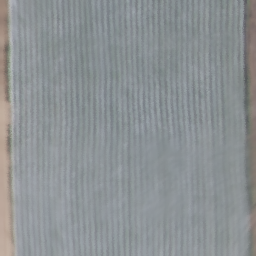}&
\includegraphics[width=0.3\linewidth]{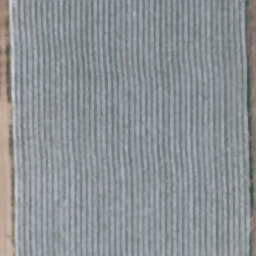}
\\[1pt]
\small GT & \small input sample&w/o ZM
\\[1pt]
\includegraphics[width=0.3\linewidth]{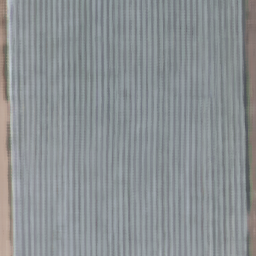}&
\includegraphics[width=0.3\linewidth]{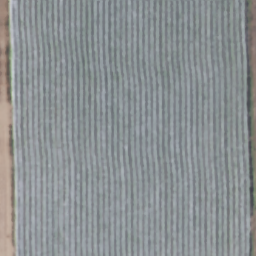}&
\includegraphics[width=0.3\linewidth]{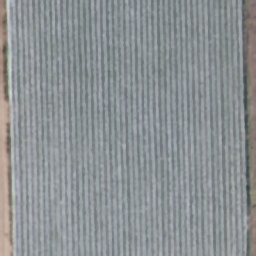}
\\[1pt]
\small w/o hashgrid PE & \small w blur ZM & \small w tilt ZM (ours)
\end{tabular}
\caption{Qualitative comparison of the ablation study on our synthetic dataset under Middle turbulence. Our full model with tilt zero temporal mean (ZM) constraint and hashgrid PE (third column) recovers a sharp image with faithful texture and geometry.}
\label{fig:ablation}
\end{figure}

\begin{table}[!t]
\centering
\footnotesize
\caption{Turbulence parameters used to synthesize on LandUse dataset~\cite{landuse}. For each scene, propagation distance $L$ and aperture $A$ are sampled uniformly, and $D/r_0$ is drawn from a discrete set. The setting follows TMFS~\cite{tmfs}.}
\label{tab:dataset_params}
\setlength{\tabcolsep}{6pt}
\renewcommand{\arraystretch}{1.15}
\begin{tabular}{@{}lccc@{}}
\toprule
Level  & $L$ [m] & $A$ [m] & $D/r_0$ \\
\midrule
Weak    & $\mathcal{U}(100, 400)$   & $\mathcal{U}(0.001, 0.005)$ & $\{0.5,\,1.0,\,1.5\}$ \\
Middle  & $\mathcal{U}(400, 800)$   & $\mathcal{U}(0.04, 0.1)$    & $\{1.0,\,1.5,\,2.0\}$ \\
Strong  & $\mathcal{U}(800, 1500)$  & $\mathcal{U}(0.1, 0.2)$     & $\{1.5,\,2.0,\,3.0\}$ \\
\bottomrule
\end{tabular}
\end{table}

\subsubsection{Datasets}
For quantitative evaluation, we use two synthetic benchmarks. The first is our own benchmark synthesized on the UCMerced LandUse dataset~\cite{landuse} using the P2S simulator~\cite{mao2021} following the settings of prior work~\cite{tmfs}, with the turbulence parameters listed in Table~\ref{tab:dataset_params}. The second is the publicly available \texttt{turb\_syn\_static} dataset~\cite{tmt}.  To evaluate robustness across turbulence strengths, we use the same random seed across the three strength levels (Weak, Middle, Strong). For qualitative evaluation, we additionally use the OTIS~\cite{otis} and RLR-AT~\cite{rlrat} real-world datasets. From each dataset, we use the first $N = 20$ frames ascending order of frame index.

\subsubsection{Comparison methods}
We compare with supervised and unsupervised methods whose source code and pretrained weights are publicly available: TMT~\cite{tmt} and DATUM~\cite{datum} as supervised methods, and CLEAR~\cite{clear} and NeRT~\cite{nert} as unsupervised methods. TMFS~\cite{tmfs}, the method most closely related to ours, is excluded from the comparison because its source code is not publicly available. All methods are evaluated on an NVIDIA L40S GPU with 48\,GB of memory. To keep the evaluation setting consistent across methods, we use the original image sizes provided in each dataset. For NeRT, we use the input images at their native spatial resolution and reconstruct the output at the same resolution (1× scale). NeRT runs out of memory (OOM) beyond a certain image size; we therefore omit the entries for datasets on which NeRT triggers OOM.

\begin{figure*}[t]
\centering
\begin{tabular}{cc}
\begin{subfigure}[t]{0.49\linewidth}
  \centering
  \setlength{\tabcolsep}{1pt}
  \renewcommand{\arraystretch}{0}
  \begin{tabular}{@{}cccc@{}}
    \includegraphics[width=0.24\linewidth]{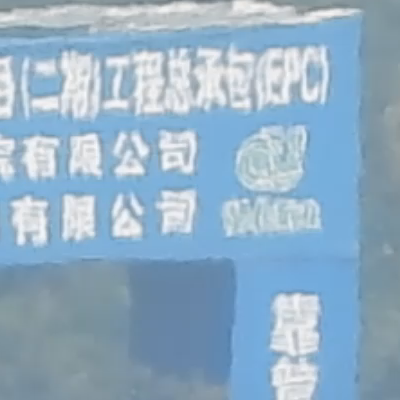} &
    \includegraphics[width=0.24\linewidth]{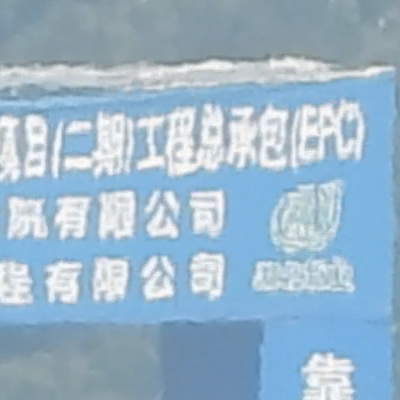} &
    \includegraphics[width=0.24\linewidth]{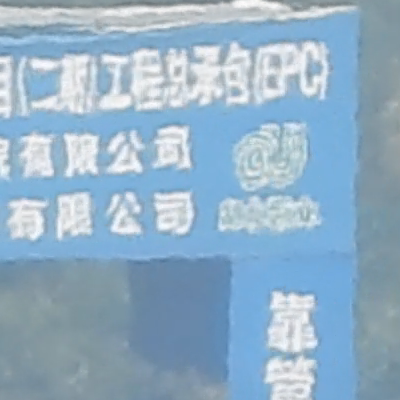} &
    \includegraphics[width=0.24\linewidth]{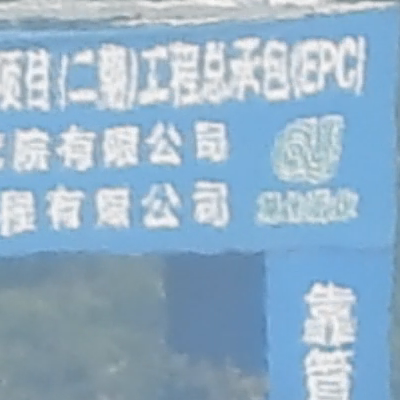} \\[1pt]
    \includegraphics[width=0.24\linewidth]{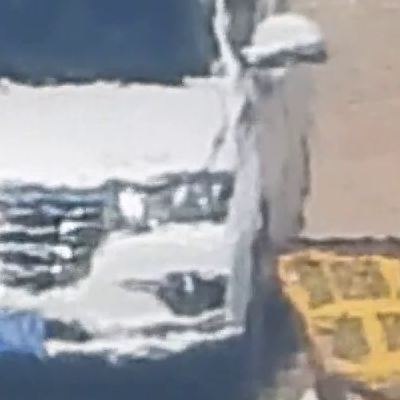} &
    \includegraphics[width=0.24\linewidth]{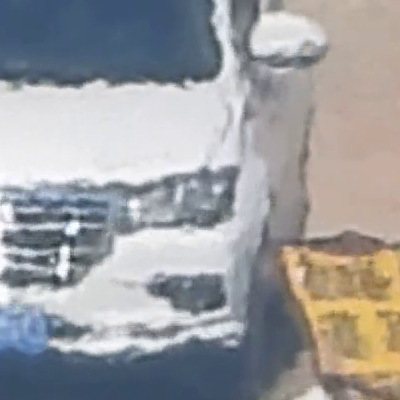} &
    \includegraphics[width=0.24\linewidth]{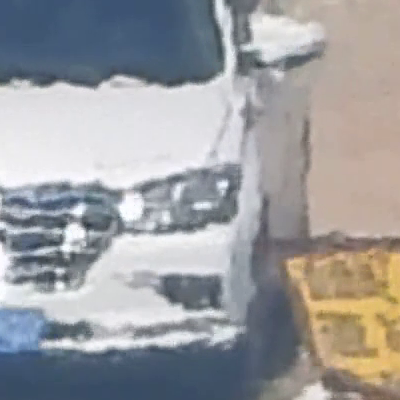} &
    \includegraphics[width=0.24\linewidth]{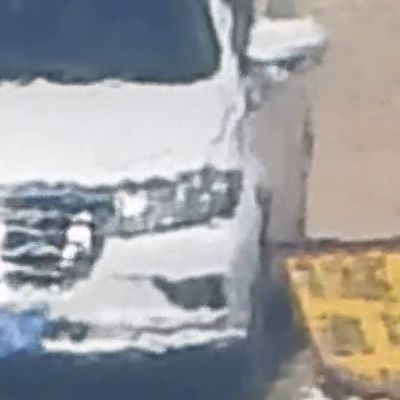}\\[1pt]
    \multicolumn{4}{c}{\vphantom{Examples of input Images}}
  \end{tabular}
  \caption{Four representative frames selected from the 20 input frames.}
\end{subfigure}
&
\begin{subfigure}[t]{0.49\linewidth}
  \centering
  \setlength{\tabcolsep}{1pt}
  \renewcommand{\arraystretch}{0}
  \begin{tabular}{@{}cccc@{}}
    \includegraphics[width=0.24\linewidth]{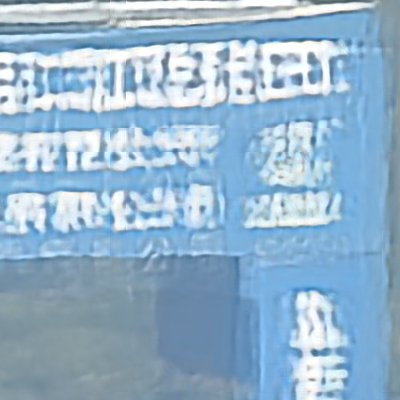} &
    \includegraphics[width=0.24\linewidth]{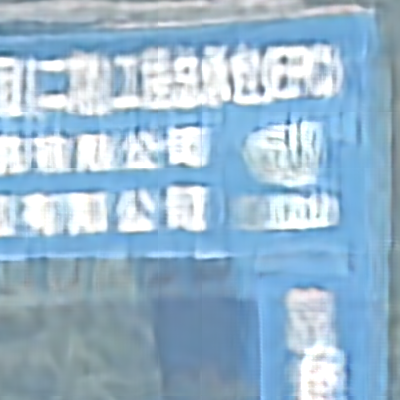} &
    \includegraphics[width=0.24\linewidth]{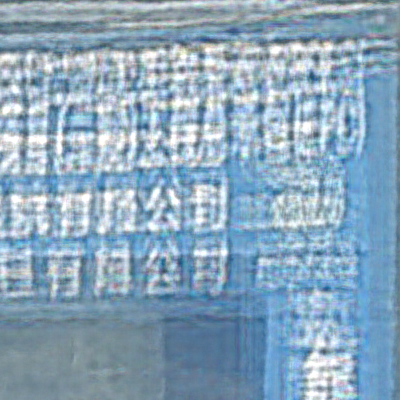} &
    \includegraphics[width=0.24\linewidth]{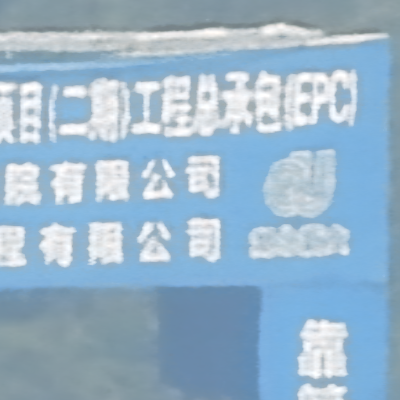}
    \\[1pt]
    \includegraphics[width=0.24\linewidth]{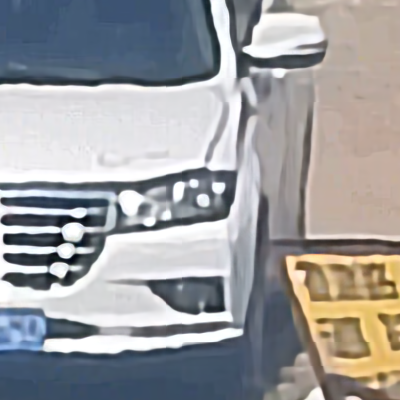} &
    \includegraphics[width=0.24\linewidth]{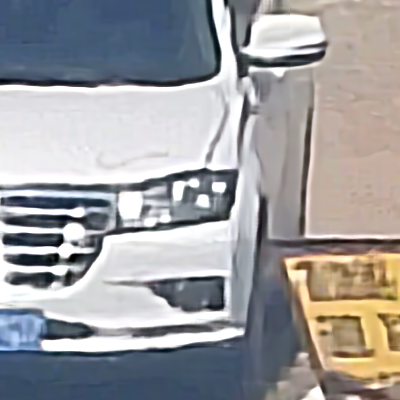} &
    \includegraphics[width=0.24\linewidth]{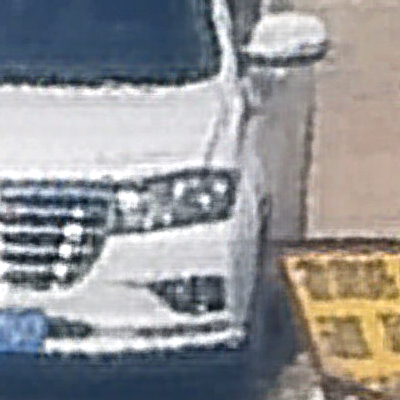} &
    \includegraphics[width=0.24\linewidth]{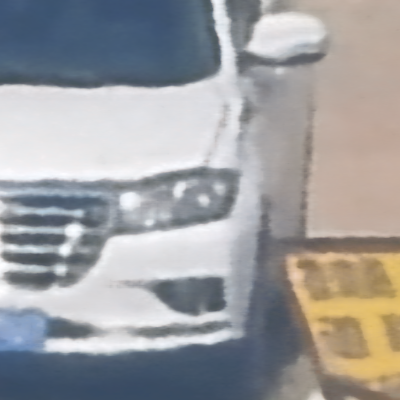}\\[1pt]
     TMT\cite{tmt}&DATUM\cite{datum}&CLEAR\cite{clear}&TriINR (ours)\\
  \end{tabular}
  \caption{Comparison of reconstructed images obtained from all 20 input frames. }
\end{subfigure}
\end{tabular}
\caption{Qualitative comparison on a real-world turbulence-degraded scene in the real-world RLR-AT dataset~\cite{rlrat}. The images are cropped for display; full-sized results are shown in the supplementary material.}
\label{fig:qual_fhd}
\end{figure*}


\subsection{Comparison results}

Table~\ref{tab:quantitative} reports the quantitative results on the \texttt{turb\_syn\_static} dataset and our synthetic dataset with three turbulence conditions. On the Weak condition, our method achieves the best PSNR and SSIM among all methods. Under the Middle and Strong conditions, our method is competitive among supervised methods and achieves the best or competitive results among unsupervised methods. On \texttt{turb\_syn\_static}, our method achieves the best or competitive PSNR among unsupervised methods. On this dataset, TMT achieves the highest accuracy because the dataset is in-domain to its training distribution. Overall, our method delivers the most consistent performance across the four benchmarks.

The qualitative results on our synthetic dataset in Figure~\ref{fig:qual_landuse} support this observation. Figure~\ref{fig:qual_otis} also shows qualitative results on the OTIS dataset. Our method recovers the pattern with the highest fidelity among all compared methods. The supervised baselines either produce visible artifacts or fail to restore a sufficient area, whereas our method yields a stable restoration. Results on additional OTIS patterns are provided in the supplementary material. 

Figure~\ref{fig:qual_fhd} presents results on RLR-AT dataset: panel (a) shows four sampled input frames, and panel (b) shows the results. For the bottom-row scene (car), all methods produce stable restoration. In contrast, the top-row scene contains large positional displacement in addition to turbulence, and methods other than ours produce strong artifacts. This shows that our method is robust not only to atmospheric turbulence but also to whole-image displacement.

\subsection{Ablation study}
\label{sec:ablation}
To understand which design choices contribute to this behavior, we ablate the two key components of TriINR: the zero-temporal-mean (ZM) constraint on the Tilt INR and the Blur INR, and the hashgrid positional encoding for the Image INR. Table~\ref{tab:ablation} summarizes the quantitative results, and Fig.~\ref{fig:ablation} shows the corresponding qualitative comparison. 

The ZM ablation investigates the effect of ZM on the tilt and on the blur separately. The bottom three rows of Table~\ref{tab:ablation} show that enforcing ZM on the tilt alone improves both PSNR and SSIM. Figure~\ref{fig:ablation} confirms this: without any ZM constraint (w/o ZM), residual tilt-induced geometric distortion remains in the restored image, whereas with ZM on the tilt (w tilt ZM) the distortion is removed. However, applying ZM to both the tilt and the blur yields slightly lower PSNR and SSIM than applying ZM only to the tilt. This result appears to contradict the underlying physical model. This supports the analysis in Section~\ref{sec:blur_zm}: the reduced sample size in the second optimization stage prevents the sample mean of the blur's Zernike coefficients from converging to the theoretical zero expectation. 

The hashgrid PE ablation examines the choice of positional encoding for the Image INR. Without the hashgrid PE, both PSNR and SSIM drop noticeably. The qualitative comparison (Fig.~\ref{fig:ablation}, w/o hashgrid) also shows that switching from the hashgrid PE to the alternative PE degrades the recovered texture fidelity. 

\begin{table}[t]
\centering
\small
\caption{Ablation study on our LandUse-based synthetic dataset under Middle turbulence.
Best values in bold. Our full model (last row) uses tilt zero-mean (ZM) without blur zero-mean, with hashgrid positional encoding for the image INR.}
\label{tab:ablation}
\begin{tabular}{cccrr}
\toprule
Tilt ZM & Blur ZM & Hashgrid PE & PSNR $\uparrow$ & SSIM $\uparrow$ \\
\midrule
\checkmark & \checkmark & -          & 23.94 & 0.605 \\
-          & -          & \checkmark & 23.94 & 0.643 \\
\checkmark & \checkmark & \checkmark & 25.54 & 0.715 \\
\checkmark & -          & \checkmark & \textbf{25.74} & \textbf{0.725} \\
\bottomrule
\end{tabular}
\end{table}

\subsection{Limitations}
\label{sec:limitations}
TriINR has two limitations that follow directly from our design choices. First, our method optimizes each sequence at test time and is therefore not real-time: restoring one sequence of $20$ images at $256 \times 256$ resolution takes around $11$ minutes on an NVIDIA L40S GPU, which is faster than the INR-based prior work NeRT (around $15$ minutes on the same GPU) but still far from the throughput of feed-forward supervised methods. Second, under Strong turbulence our method is outperformed by supervised methods on PSNR/SSIM (Table~\ref{tab:quantitative}). Supervised methods can fill in plausible textures using priors learned from paired data, whereas our unsupervised formulation has no such prior.

\section{Conclusion}

We presented TriINR, an unsupervised atmospheric TM framework. Existing INR-based TM methods suffer from frame-dependent degradation representations. We reformulated the sharp image, the tilt, and the blur as continuous INRs. This yields frame-independent parameter scaling and memory-efficient optimization. Moreover, we incorporated physical and statistical properties of atmospheric turbulence as constraints in the architecture.  We embed the zero-temporal-mean constraint on tilt directly into the Tilt INR architecture, and let the Blur INR predict per-pixel PSFs. In experiments, TriINR outperformed prior unsupervised INR-based methods and achieved robust reconstruction across varying turbulence strengths. For future work, we plan to accelerate the optimization for faster convergence and to incorporate priors as regularization to better handle strong turbulence.

\label{sec:conclusion}

{
    \small
    \bibliographystyle{ieeenat_fullname}
    \bibliography{main}
}

\newpage
\clearpage
\setcounter{page}{1}
\maketitlesupplementary

In this supplementary material, we first present additional qualitative results that were omitted from the main paper due to space constraints. Second, we show more detailed architectural comparison against prior INR-based turbulence mitigation (TM) methods. We then discuss the representational capacity of our INRs with respect to the number of input frames. Finally, we then provide visualization of our two-stage optimization procedure. 

\section{Additional Qualitative Results}
\label{sec:supp_qualitative}

Figure~\ref{fig:qual_sup} presents additional qualitative comparisons on our LandUse-based synthetic dataset. Figure~\ref{fig:qual_otis_odd} and Figure~\ref{fig:qual_otis_even} show comparisons on the OTIS dataset~\cite{otis} across a wider range of patterns. Across all these scenes, TriINR consistently produces robust restorations with faithful texture and geometry. 

\begin{figure*}[t]
\centering
\begin{tabular}{cc}
    \includegraphics[width=0.45\linewidth]{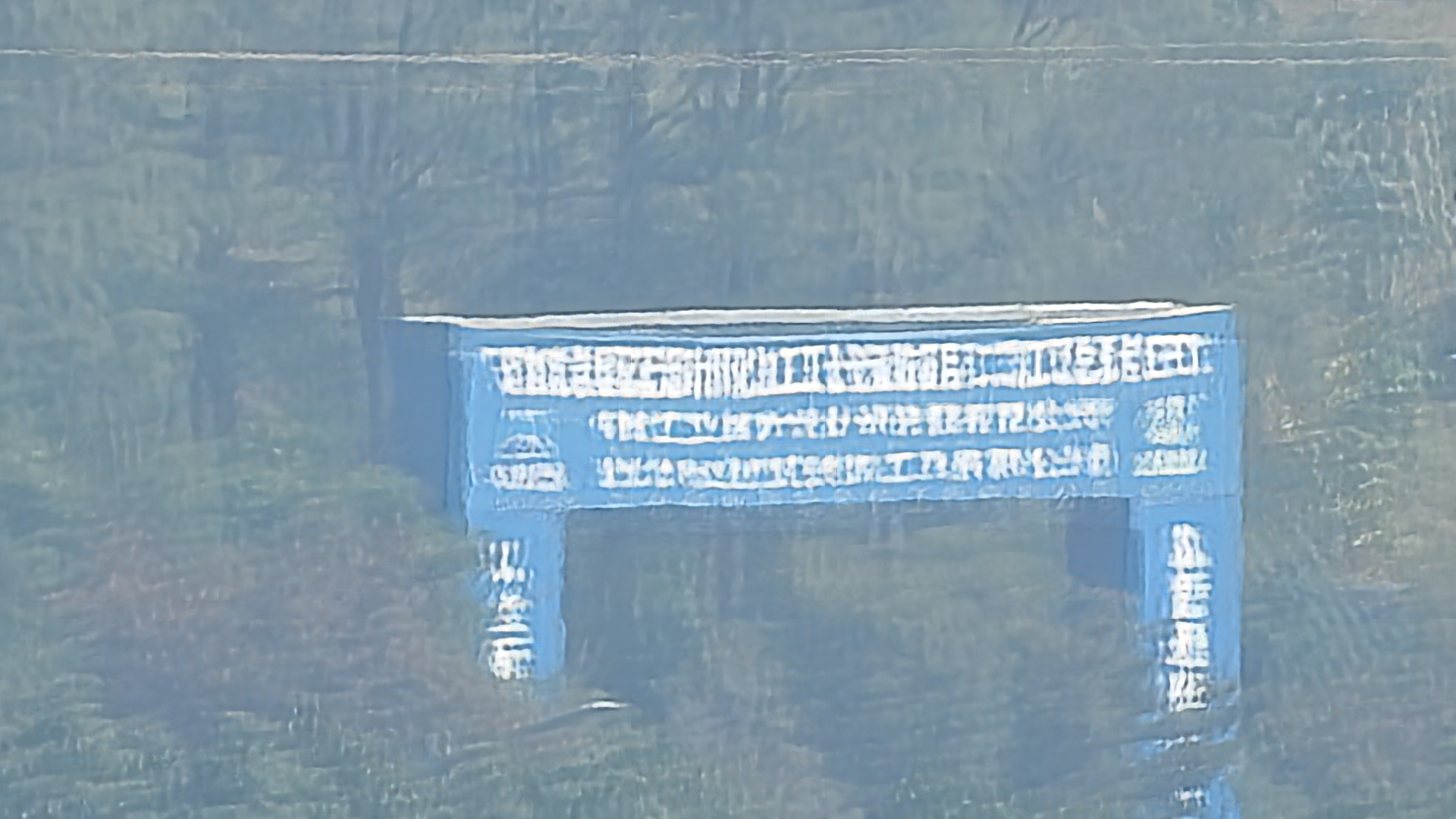} &
    \includegraphics[width=0.45\linewidth]{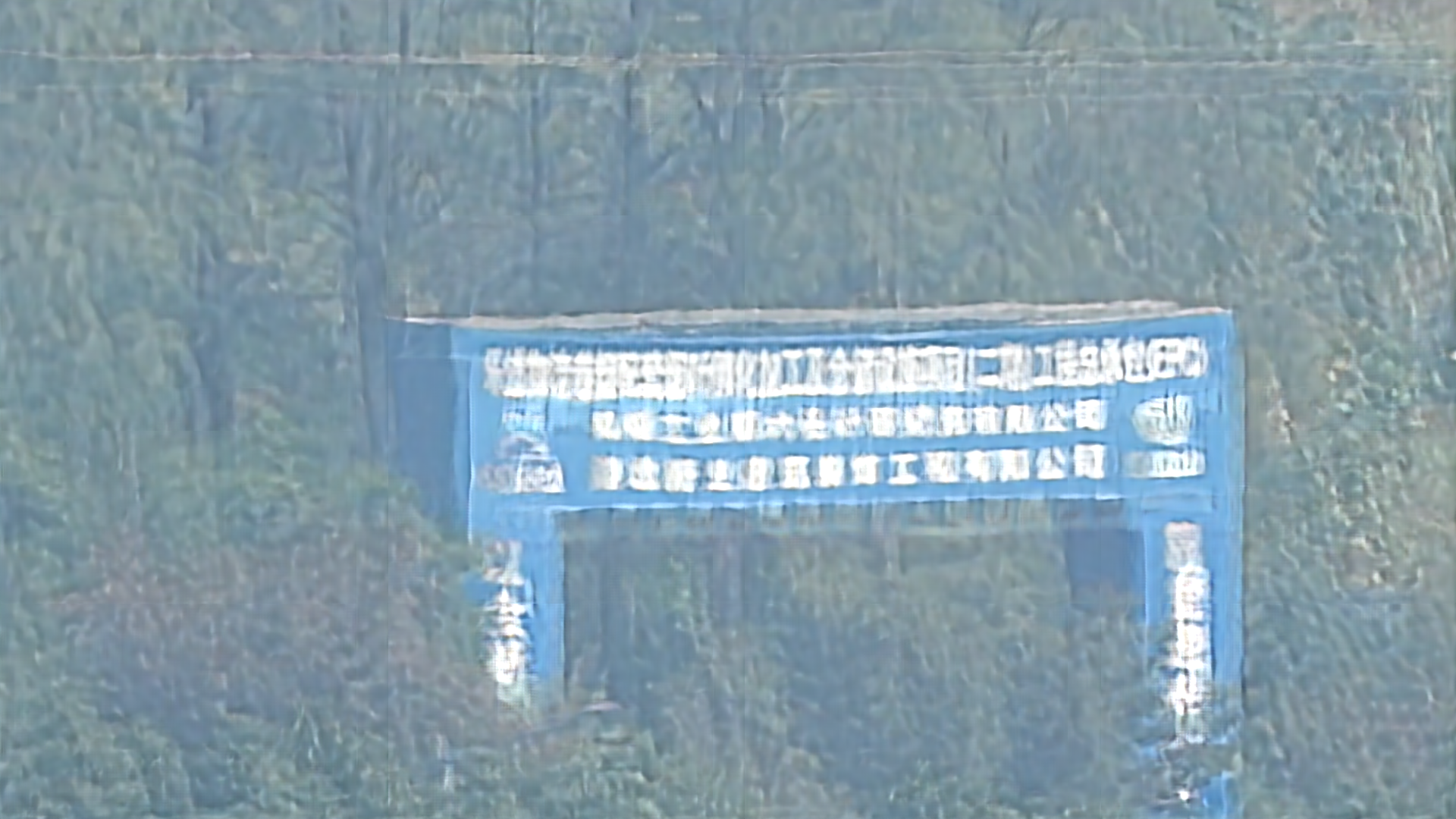}\\
    TMT \cite{tmt} & DATUM \cite{datum}\\
    \includegraphics[width=0.45\linewidth]{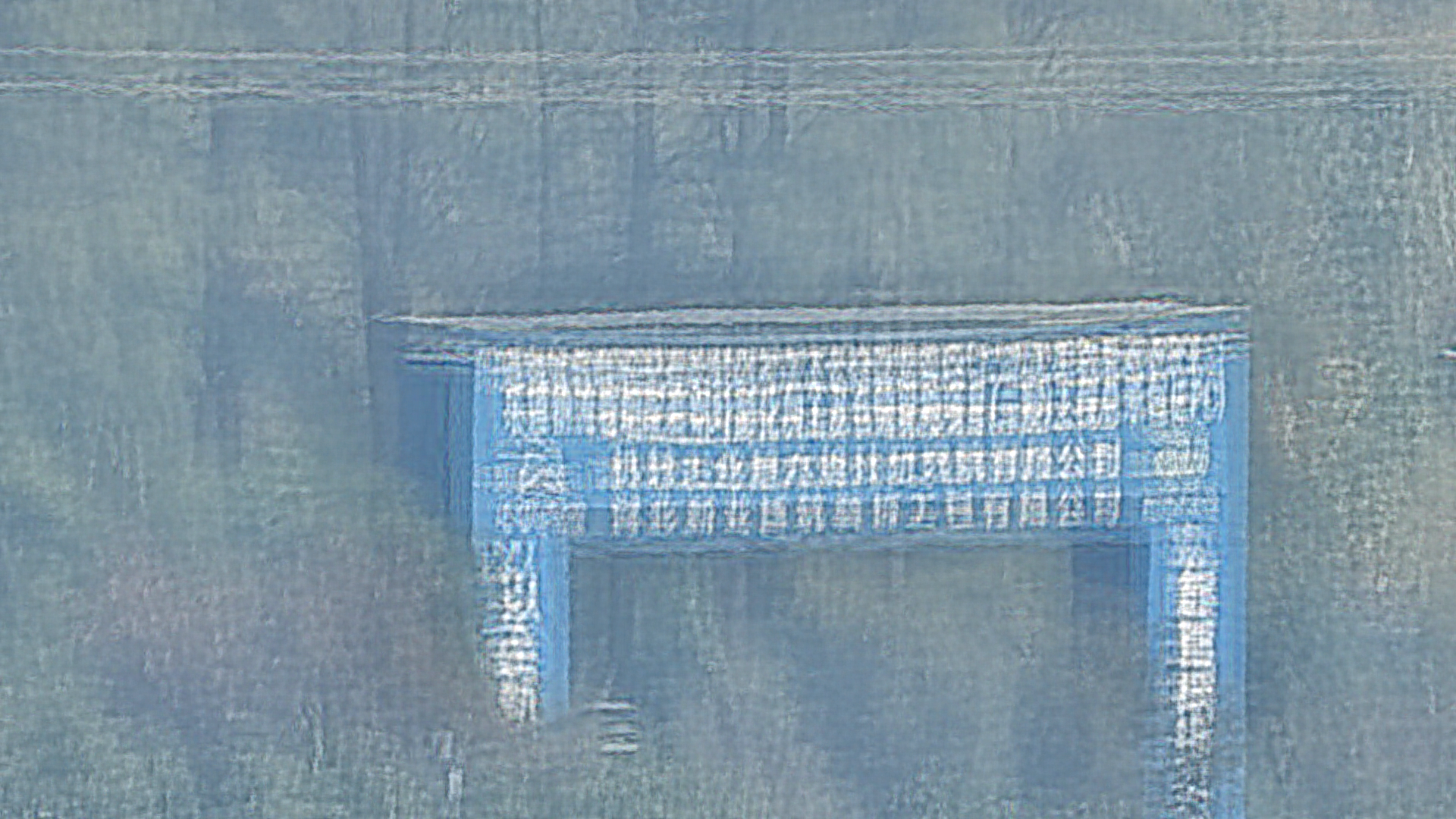} &
    \includegraphics[width=0.45\linewidth]{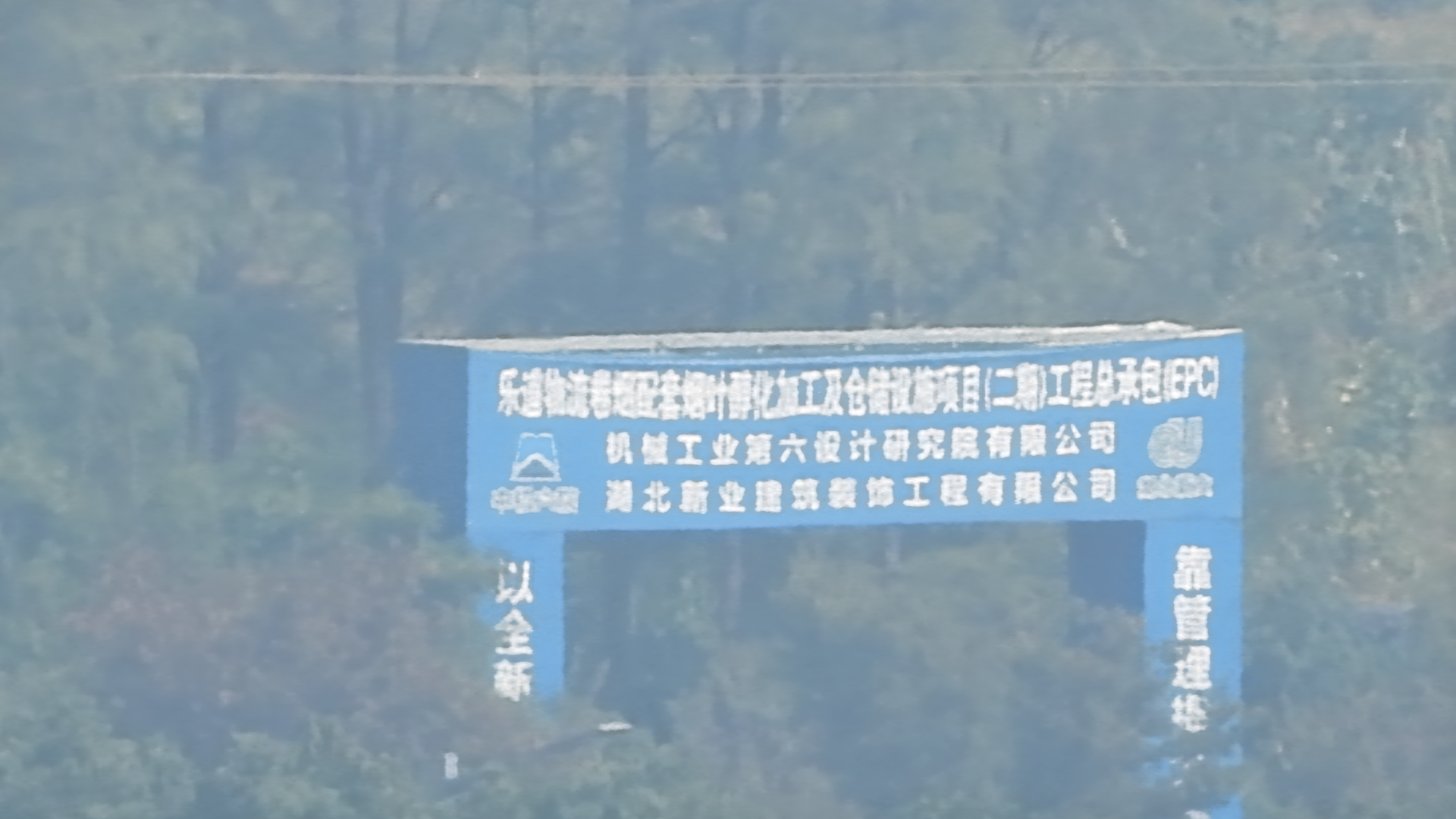}\\
    CLEAR \cite{clear} & TriINR (ours)\\
    \includegraphics[width=0.45\linewidth]{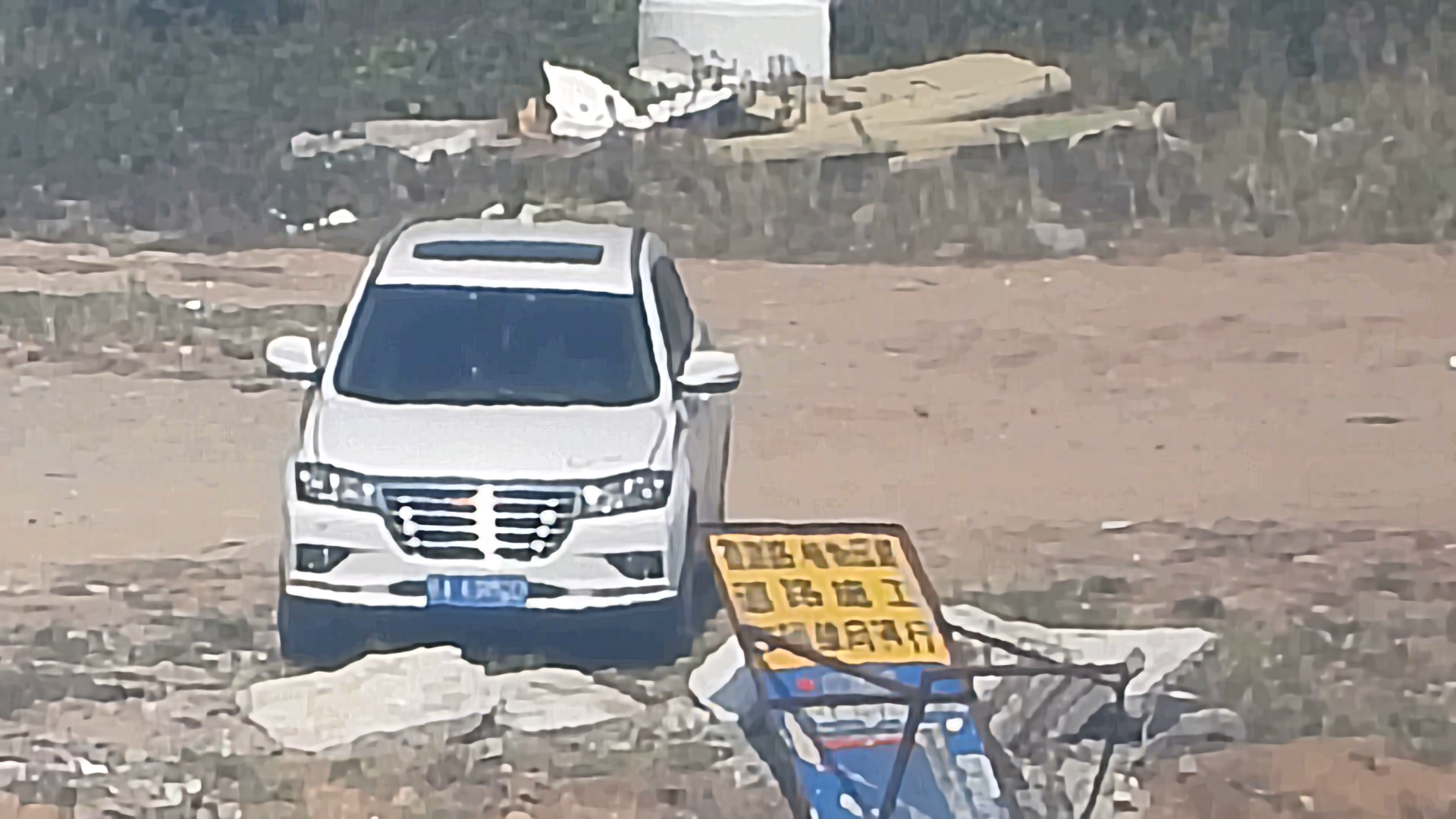} &
    \includegraphics[width=0.45\linewidth]{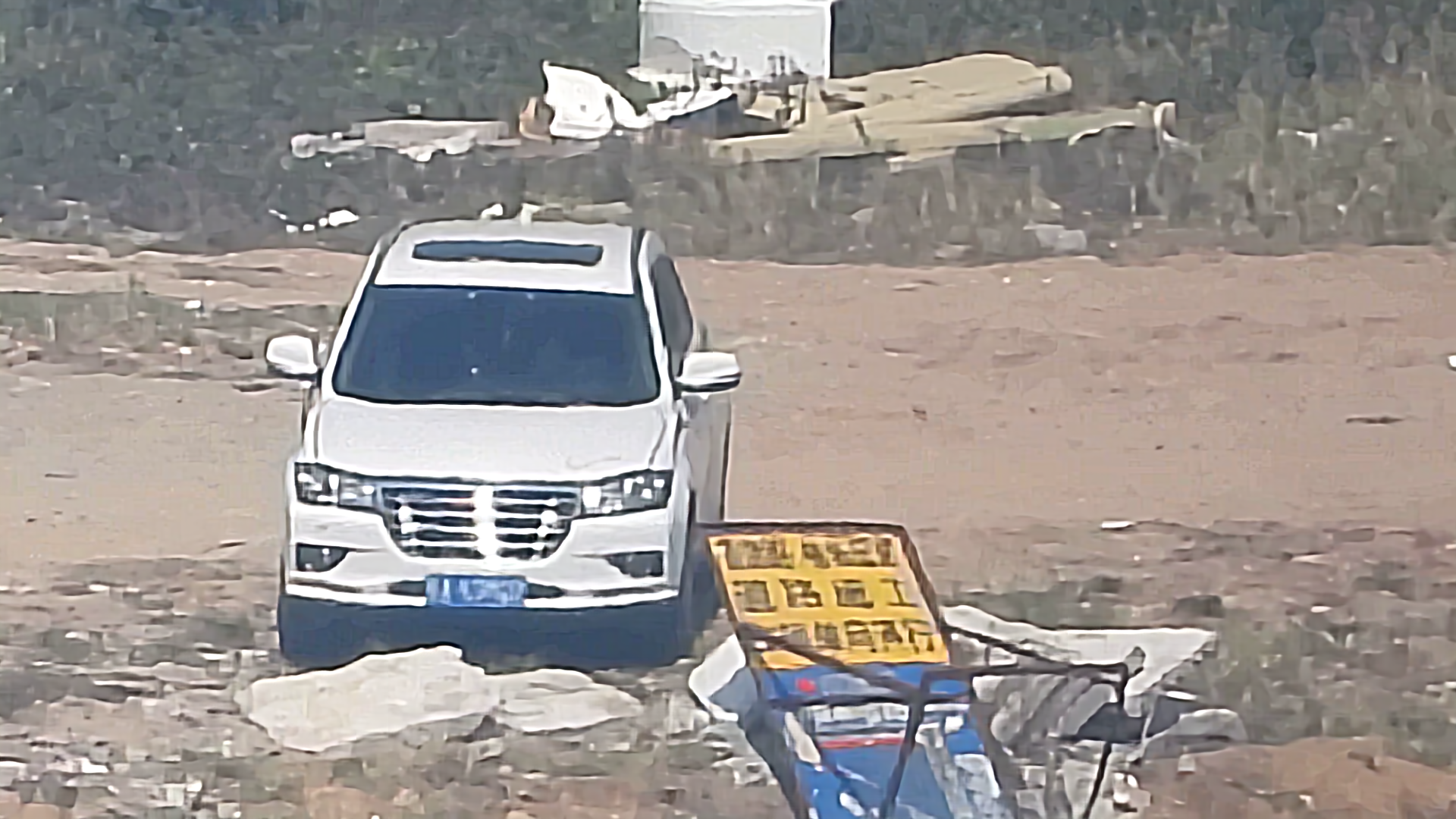}\\
    TMT \cite{tmt} & DATUM \cite{datum}\\
    \includegraphics[width=0.45\linewidth]{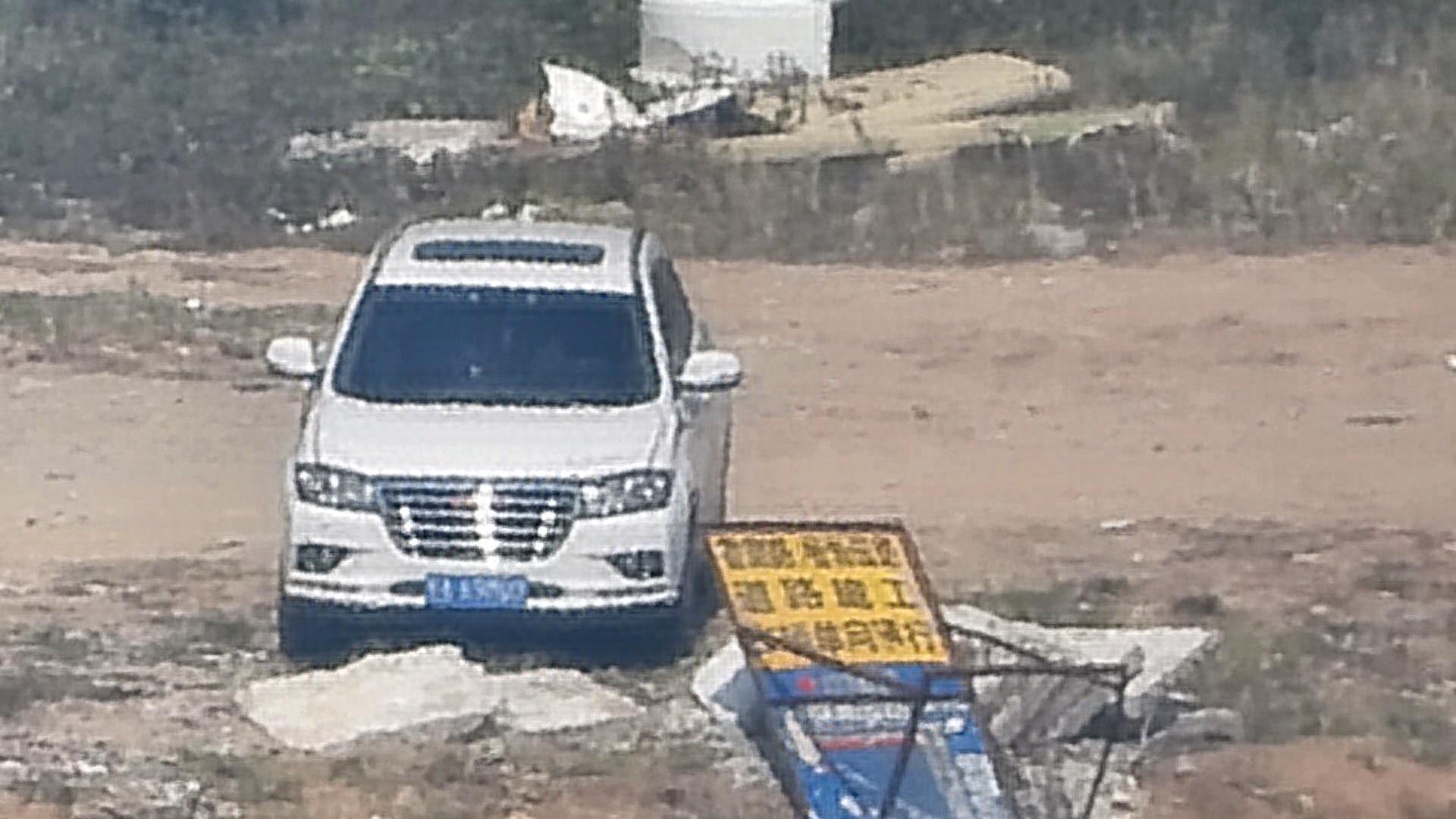} &
    \includegraphics[width=0.45\linewidth]{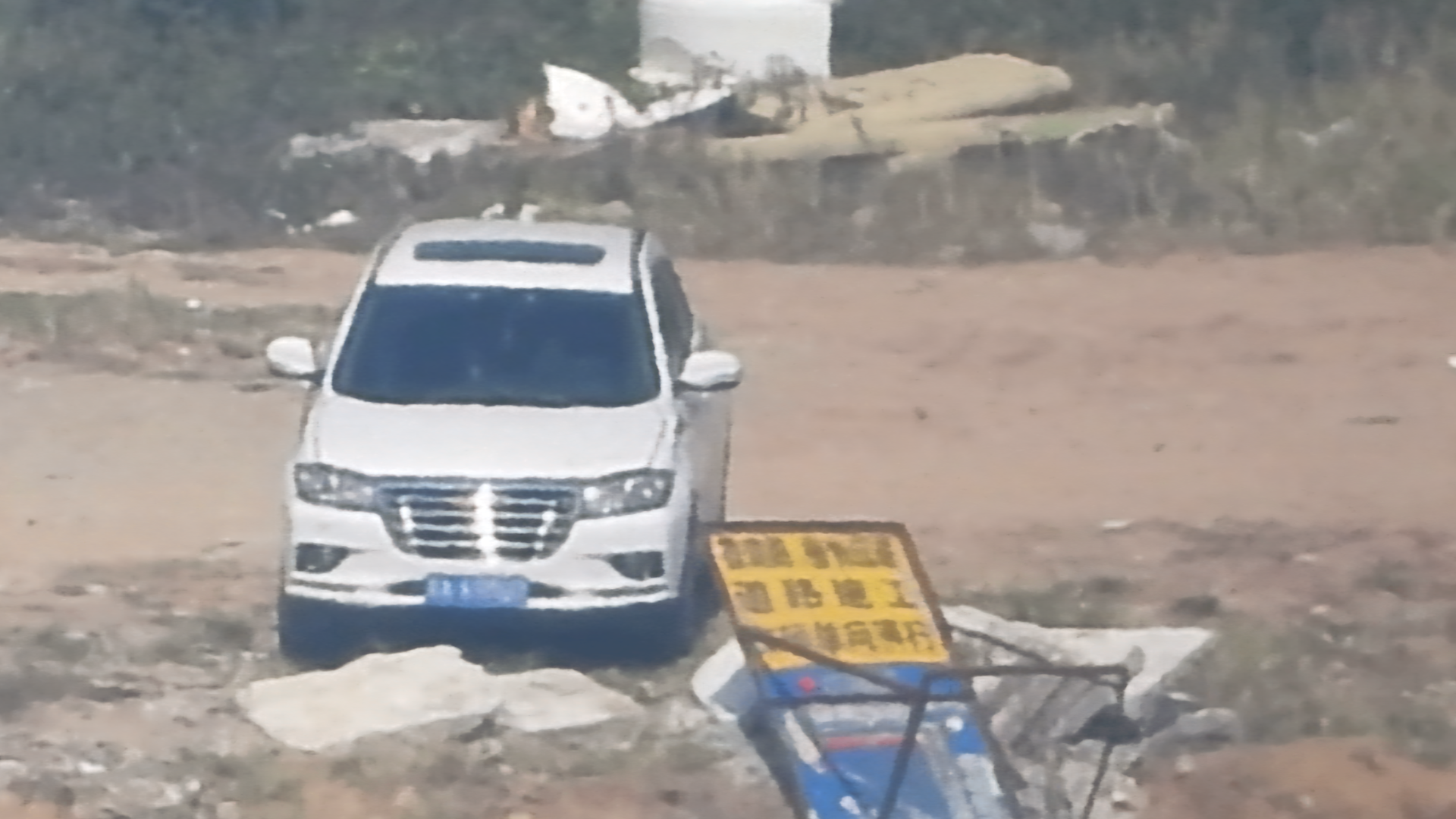}\\
    CLEAR \cite{clear} & TriINR (ours)
\end{tabular}
\caption{FHD comparison results of RLR-AT \cite{rlrat}}
\label{fig:qual_fhd_full}
\end{figure*}

Figure~\ref{fig:qual_fhd_full} shows the full-sized restoration results on the RLR-AT~\cite{rlrat} FHD scene that were cropped for display in the main paper (Figure~\ref{fig:qual_fhd}). The full-sized comparison confirms that TriINR remains effective on high-resolution inputs and is robust to whole-image displacement in addition to atmospheric turbulence.

\section{Detailed Architectural Comparison with INR-based Methods}
\label{sec:supp_arch}

Table~\ref{tab:param_compare2} provides a more detailed comparison of the architectures and parameter counts of INR-based turbulence mitigation methods. 
NeRT~\cite{nert} instantiates a per-frame CNN grid deformer for the tilt and stores per-frame Zernike-vector tensors for the blur, so both the tilt and the blur parameters grow linearly with the number of input frames.
TMFS~\cite{tmfs} additionally scales its tilt and blur parameters with the image resolution, and therefore has to tile the input into $256 \times 256$ patches.
In contrast, TriINR represents all three components with fixed-size MLPs whose parameter count is independent of both the number of input frames and the image resolution.

\section{Effect of the Number of Input Frames}
\label{sec:supp_n_ablation}

Although the parameter count of our INRs is independent of $N$ (Section~3.3.4 of the main paper), the representational capacity of the INRs at a fixed parameter budget bounds the spatio-temporal complexity they can encode.
As shown in Table~\ref{tab:n_ablation}, restoration accuracy improves as $N$ grows up to $N{=}20$, but PSNR and SSIM decline for larger $N$.
This indicates that the tilt and blur variations of a longer sequence cannot all be captured under the current network sizes.
Scaling up the MLP widths or the number of hashgrid levels would allow representing sequences with more frames, at the cost of additional parameters and memory.

\begin{table}[t]
\centering
\footnotesize
\caption{Effect of the input frame count $N$ on our LandUse-based synthetic dataset under Middle turbulence. PSNR and SSIM are averaged over 105 scenes (5 initial scenes per class $\times$ 21 classes). Best value in each row is in \textbf{bold}.}
\label{tab:n_ablation}
\setlength{\tabcolsep}{6pt}
\renewcommand{\arraystretch}{1.15}
\begin{tabular}{@{}l c c c c c@{}}
\toprule
$N$              & 5     & 10    & 20              & 40    & 80    \\
\midrule
PSNR $\uparrow$  & 24.83 & 25.55 & \textbf{25.77}  & 25.50 & 25.14 \\
SSIM $\uparrow$  & 0.690 & 0.722 & \textbf{0.725}  & 0.709 & 0.688 \\
\bottomrule
\end{tabular}
\end{table}
\begin{figure}[t]
\centering
\setlength{\tabcolsep}{1pt}
\centering
\setlength{\tabcolsep}{1pt}
\renewcommand{\arraystretch}{0}
\begin{tabular}{@{}cc@{}}
\includegraphics[width=0.31\linewidth]{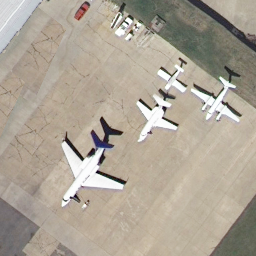}&
\includegraphics[width=0.31\linewidth]{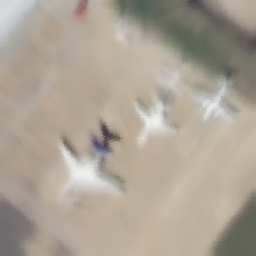}\\[1pt] 
\small GT&\small Synthesized example\\[1pt]
\includegraphics[width=0.31\linewidth]{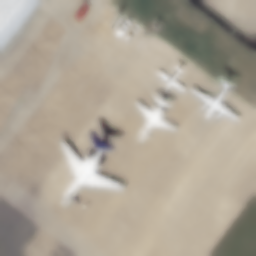}&
\includegraphics[width=0.31\linewidth]{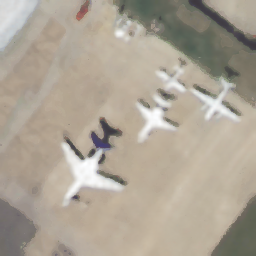}\\[1pt]
\small After 1st optimized result&\small Final image 
\end{tabular}
\caption{Results of the two-stage optimization on a synthesized image from LandUse dataset. Top row: the ground-truth sharp image (left) and one synthesized input frame (right). Bottom row: the Image INR output after stage 1 (left) and after stage 2 (right).}
\label{fig:opt_results_small}
\end{figure}

\section{Effect of the Two-Stage Optimization}
\label{sec:supp_two_stage}

Figure~\ref{fig:opt_results_small} visualizes the intermediate result after stage~1 and the final result after stage~2 on a synthesized sequence.
Stage~1 optimizes only the Image INR and the Tilt INR with a frozen delta PSF, which is already sufficient to remove most of the geometric distortion caused by atmospheric tilt.
Stage~2 then unfreezes the Blur INR and jointly optimizes all three INRs on the full forward model, further sharpening the image by capturing the residual spatially varying blur.
The recovered tilt map and per-pixel PSFs closely reproduce the ground-truth structures, confirming that our joint optimization successfully disentangles the tilt and the blur components of atmospheric turbulence.

\begin{table*}[t]
\centering
\small
\caption{Parameter and representation comparison of unsupervised INR-based turbulence mitigation methods. Unlike prior methods, whose parameter count grows with the frame count and image size, TriINR keeps a constant parameter count. PE denotes positional encoding, and GF denotes Gaussian Fourier features~\cite{fourier_features}.}
\label{tab:param_compare2}
\setlength{\tabcolsep}{4pt}
\renewcommand{\arraystretch}{1.15}
\begin{tabular}{@{}l l l l@{}}
\toprule
& NeRT~\cite{nert} & TMFS~\cite{tmfs} & \textbf{TriINR (ours)} \\
\midrule
Tilt
    & Frame-dependent INRs
    & Frame-dependent tensors
    & \textbf{Frame-independent INR} \\
representation
    & Grid deformer~\cite{ndir}, one CNN / frame
    & Noise map + correlation
    & PE (GF) + MLP, $(x,y,t)$ input \\
    & ($1{\times}1$ conv, 256ch, $\times 4$)
    & $(2{\times}2W{\times}2H + 2)$ / frame
    & MLP width ${=} 128 \times 4$ \\
\midrule
Blur
    & Frame-dependent tensors
    & Frame- \& resolution-dependent tensors
    & \textbf{Frame-independent INR} \\
representation
    & Zernike vector map
    & Noise map + correlation parameters
    & PE (GF) + MLP, $(x,y,t)$ input \\
    & $16{\times}16{\times}36$ / frame
    & $(19{\times}2{\times}2W{\times}2H + 19)$ / frame
    & MLP width ${=} 128 \times 4$ \\
\midrule
Image
    & Single INR
    & Single INR
    & Single INR \\
representation
    & PE (GF) + MLP
    & MLP (PE unspecified)
    & PE (Hashgrid~\cite{instantngp}) + MLP \\
    & MLP width ${=} 256 \times 5$
    & 4 layers (details unspecified)
    & MLP width ${=} 64 \times 4$ \\
\midrule
Parameter count scaling
    & $O(N)$
    & $O(H \times W \times N)$
    & ${O(1)}$ \\
\midrule
\#Parameters {\tiny (H=W=256, N=20)}
    & $\sim 3.8$M
    & $\sim 200$M
    & $\sim \mathbf{0.4M}$ \\
\bottomrule
\end{tabular}
\end{table*}

\begin{figure*}[t]
\centering
\setlength{\tabcolsep}{1pt}
\renewcommand{\arraystretch}{0}
\begin{tabular}{@{}c@{\hspace{2pt}}ccccccc@{}}
\adjustbox{valign=c,rotate=90}{\small \quad \quad Weak} &
\includegraphics[width=0.13\linewidth]{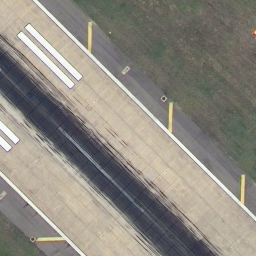}&
\includegraphics[width=0.13\linewidth]{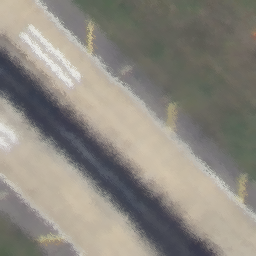}&
\includegraphics[width=0.13\linewidth]{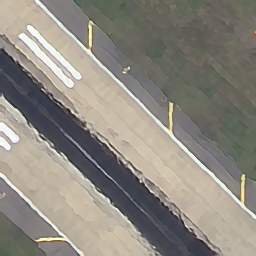}&
\includegraphics[width=0.13\linewidth]{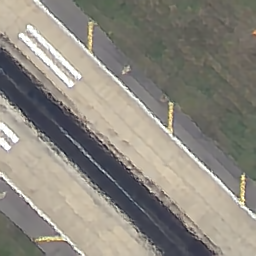}&
\includegraphics[width=0.13\linewidth]{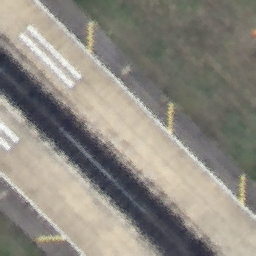}&
\includegraphics[width=0.13\linewidth]{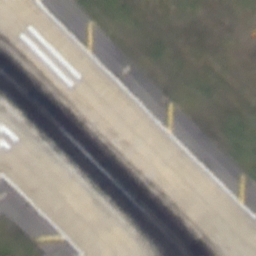}&
\includegraphics[width=0.13\linewidth]{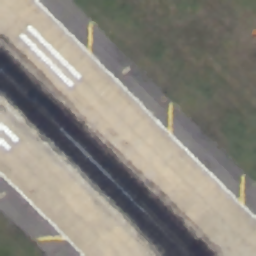}\\[1pt]
\adjustbox{valign=c,rotate=90}{\small \quad \quad Middle} &
\includegraphics[width=0.13\linewidth]{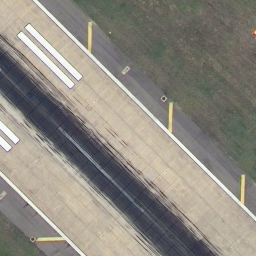}&
\includegraphics[width=0.13\linewidth]{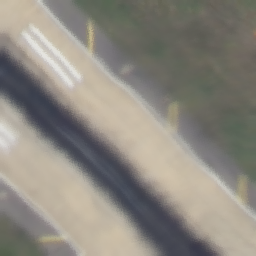}&
\includegraphics[width=0.13\linewidth]{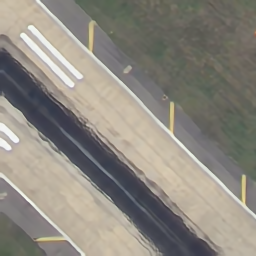}&
\includegraphics[width=0.13\linewidth]{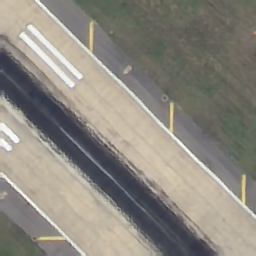}&
\includegraphics[width=0.13\linewidth]{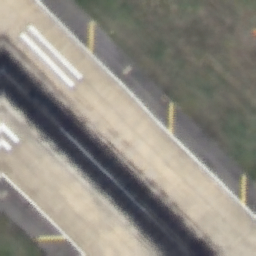}&
\includegraphics[width=0.13\linewidth]{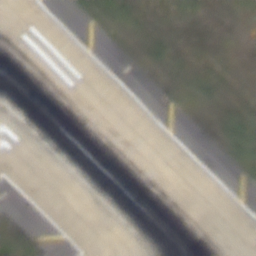}&
\includegraphics[width=0.13\linewidth]{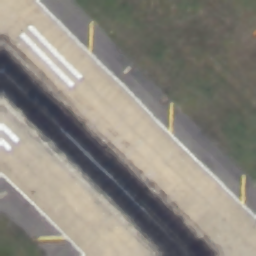}\\[1pt]
\adjustbox{valign=c,rotate=90}{\small \quad \quad Strong} &
\includegraphics[width=0.13\linewidth]{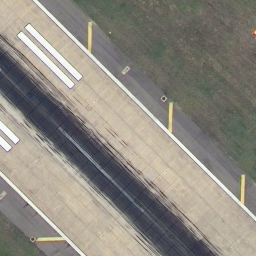}&
\includegraphics[width=0.13\linewidth]{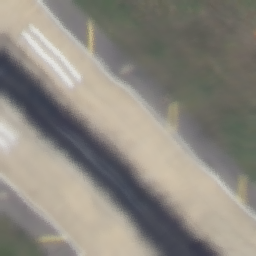}&
\includegraphics[width=0.13\linewidth]{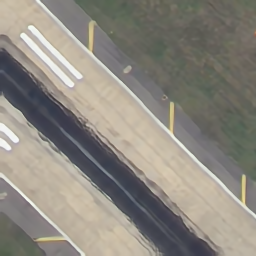}&
\includegraphics[width=0.13\linewidth]{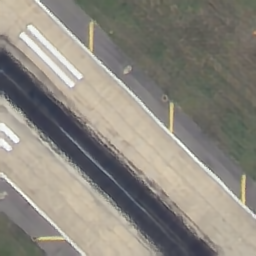}&
\includegraphics[width=0.13\linewidth]{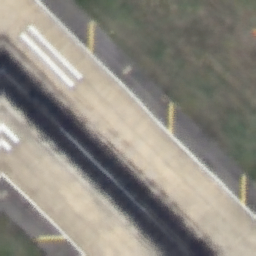}&
\includegraphics[width=0.13\linewidth]{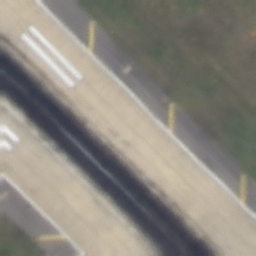}&
\includegraphics[width=0.13\linewidth]{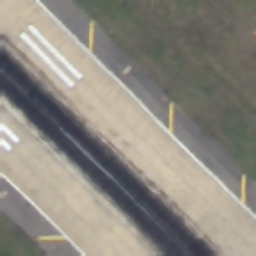}\\[5pt]
\adjustbox{valign=c,rotate=90}{\small \quad \quad Weak} &
\includegraphics[width=0.13\linewidth]{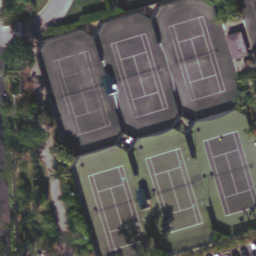}&
\includegraphics[width=0.13\linewidth]{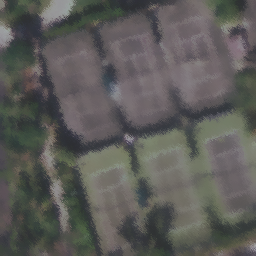}&
\includegraphics[width=0.13\linewidth]{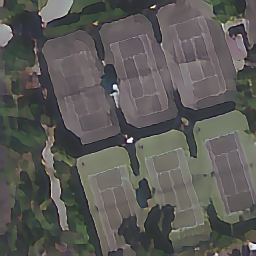}&
\includegraphics[width=0.13\linewidth]{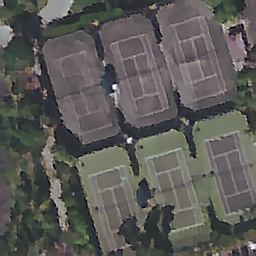}&
\includegraphics[width=0.13\linewidth]{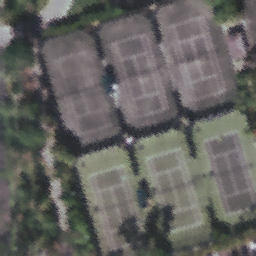}&
\includegraphics[width=0.13\linewidth]{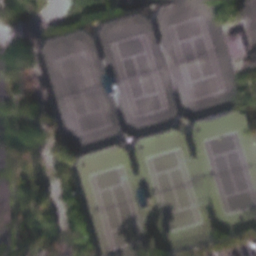}&
\includegraphics[width=0.13\linewidth]{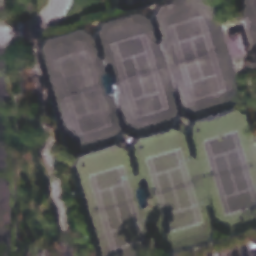}\\[1pt]
\adjustbox{valign=c,rotate=90}{\small \quad \quad Middle} &
\includegraphics[width=0.13\linewidth]{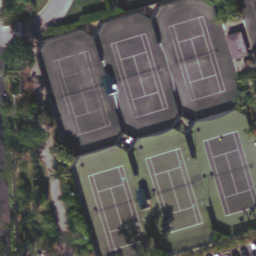}&
\includegraphics[width=0.13\linewidth]{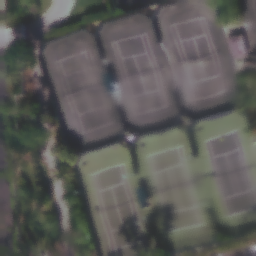}&
\includegraphics[width=0.13\linewidth]{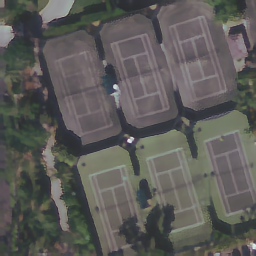}&
\includegraphics[width=0.13\linewidth]{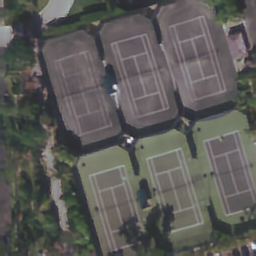}&
\includegraphics[width=0.13\linewidth]{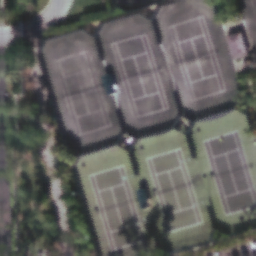}&
\includegraphics[width=0.13\linewidth]{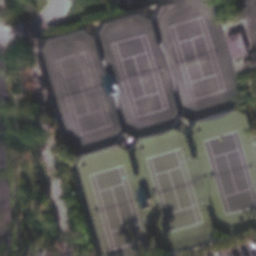}&
\includegraphics[width=0.13\linewidth]{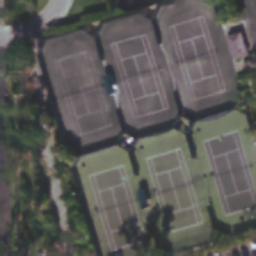}\\[1pt]
\adjustbox{valign=c,rotate=90}{\small \quad \quad Strong} &
\includegraphics[width=0.13\linewidth]{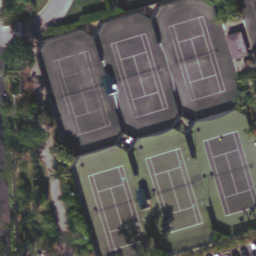}&
\includegraphics[width=0.13\linewidth]{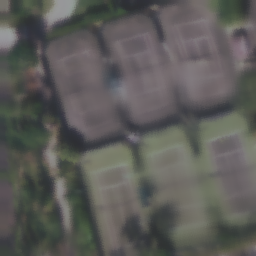}&
\includegraphics[width=0.13\linewidth]{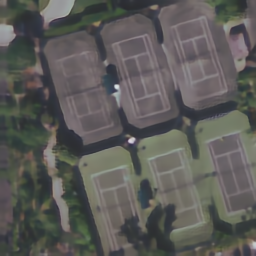}&
\includegraphics[width=0.13\linewidth]{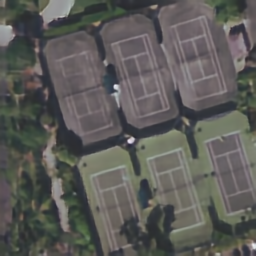}&
\includegraphics[width=0.13\linewidth]{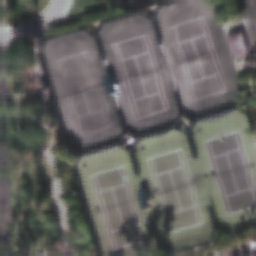}&
\includegraphics[width=0.13\linewidth]{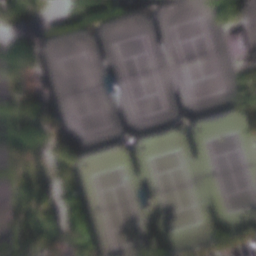}&
\includegraphics[width=0.13\linewidth]{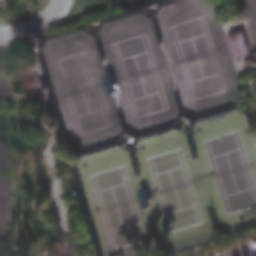}\\[1pt]

& \small GT & \small Input sample & \small TMT \cite{tmt} & \small DATUM \cite{datum} & \small CLEAR \cite{clear} & \small NeRT \cite{nert} & \small TriINR (ours) \\
\end{tabular}
\caption{Additional qualitative comparison on our LandUse-based synthetic dataset under three turbulence strengths (Weak / Middle / Strong; top to bottom), each restored from 20 input frames.}
\label{fig:qual_sup}
\end{figure*}

\begin{figure*}[t]
\centering
\setlength{\tabcolsep}{1pt}
\renewcommand{\arraystretch}{0}
\begin{tabular}{@{}c@{\hspace{2pt}}cccccc@{}}
\adjustbox{valign=c,rotate=90}{\small \quad \quad \quad Pattern1} &
\includegraphics[width=0.15\linewidth]{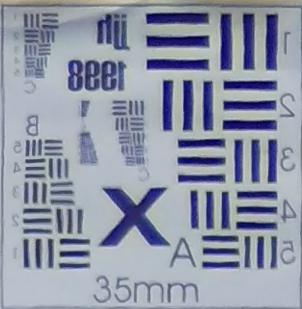}&
\includegraphics[width=0.15\linewidth]{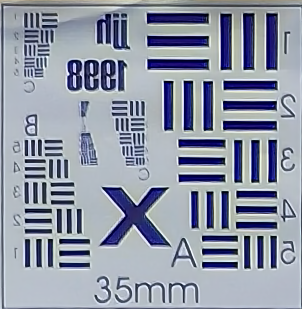}&
\includegraphics[width=0.15\linewidth]{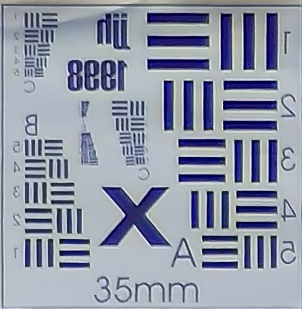}&
\includegraphics[width=0.15\linewidth]{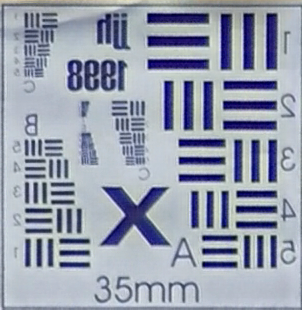}&
&
\includegraphics[width=0.15\linewidth]{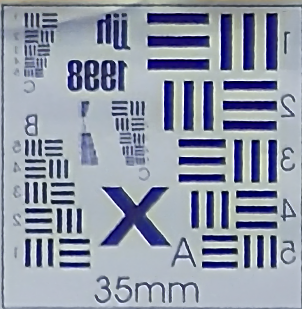}
\\[1pt]
\adjustbox{valign=c,rotate=90}{\small \quad \quad \quad Pattern3} &
\includegraphics[width=0.15\linewidth]{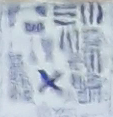}&
\includegraphics[width=0.15\linewidth]{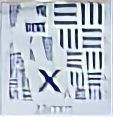}&
\includegraphics[width=0.15\linewidth]{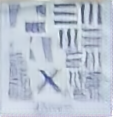}&
\includegraphics[width=0.15\linewidth]{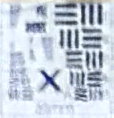}&
\includegraphics[width=0.15\linewidth]{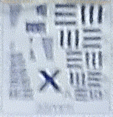}&
\includegraphics[width=0.15\linewidth]{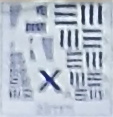}
\\[1pt]
\adjustbox{valign=c,rotate=90}{\small \quad \quad \quad Pattern5} &
\includegraphics[width=0.15\linewidth]{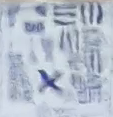}&
\includegraphics[width=0.15\linewidth]{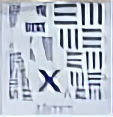}&
\includegraphics[width=0.15\linewidth]{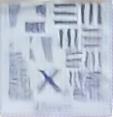}&
\includegraphics[width=0.15\linewidth]{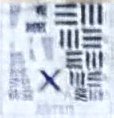}&
\includegraphics[width=0.15\linewidth]{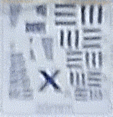}&
\includegraphics[width=0.15\linewidth]{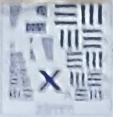}
\\[1pt]
\adjustbox{valign=c,rotate=90}{\small \quad \quad \quad Pattern7} &
\includegraphics[width=0.15\linewidth]{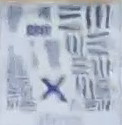}&
\includegraphics[width=0.15\linewidth]{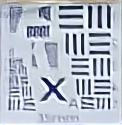}&
\includegraphics[width=0.15\linewidth]{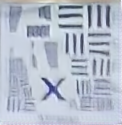}&
\includegraphics[width=0.15\linewidth]{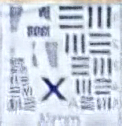}&
\includegraphics[width=0.15\linewidth]{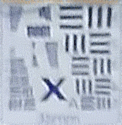}&
\includegraphics[width=0.15\linewidth]{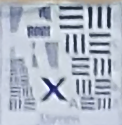}
\\[1pt]
\adjustbox{valign=c,rotate=90}{\small \quad \quad \quad Pattern9} &
\includegraphics[width=0.15\linewidth]{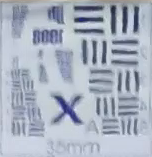}&
\includegraphics[width=0.15\linewidth]{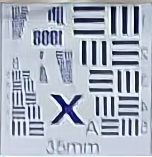}&
\includegraphics[width=0.15\linewidth]{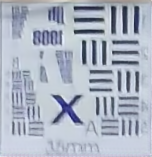}&
\includegraphics[width=0.15\linewidth]{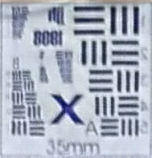}&
\includegraphics[width=0.15\linewidth]{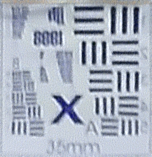}&
\includegraphics[width=0.15\linewidth]{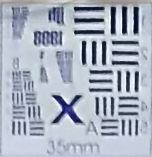}
\\[1pt]
\adjustbox{valign=c,rotate=90}{\small \quad \quad \quad Pattern11} &
\includegraphics[width=0.15\linewidth]{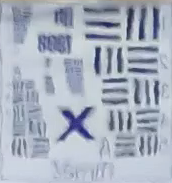}&
\includegraphics[width=0.15\linewidth]{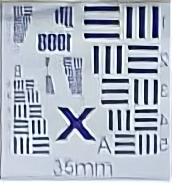}&
\includegraphics[width=0.15\linewidth]{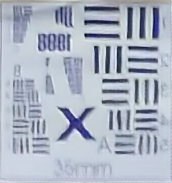}&
\includegraphics[width=0.15\linewidth]{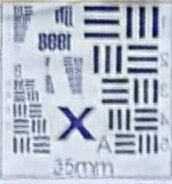}&
\includegraphics[width=0.15\linewidth]{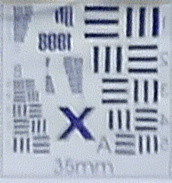}&
\includegraphics[width=0.15\linewidth]{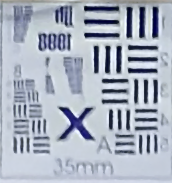}
\\[1pt]
\adjustbox{valign=c,rotate=90}{\small \quad \quad \quad Pattern13} &
\includegraphics[width=0.15\linewidth]{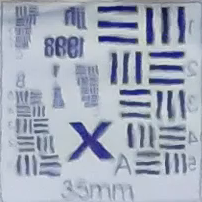}&
\includegraphics[width=0.15\linewidth]{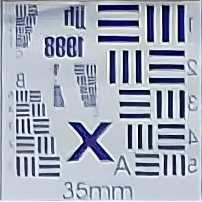}&
\includegraphics[width=0.15\linewidth]{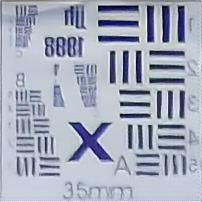}&
\includegraphics[width=0.15\linewidth]{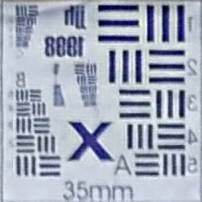}&
\includegraphics[width=0.15\linewidth]{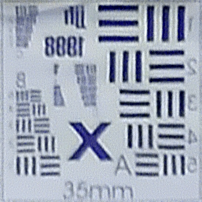}&
\includegraphics[width=0.15\linewidth]{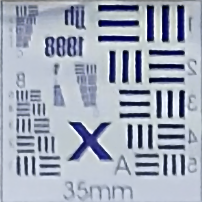}
\\[1pt]
\adjustbox{valign=c,rotate=90}{\small \quad \quad \quad Pattern15} &
\includegraphics[width=0.15\linewidth]{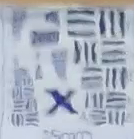}&
\includegraphics[width=0.15\linewidth]{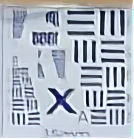}&
\includegraphics[width=0.15\linewidth]{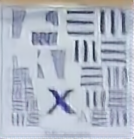}&
\includegraphics[width=0.15\linewidth]{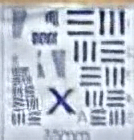}&
\includegraphics[width=0.15\linewidth]{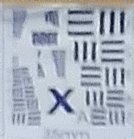}&
\includegraphics[width=0.15\linewidth]{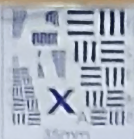}
\\[1pt]
&  \small Input sample & \small TMT \cite{tmt} & \small DATUM \cite{datum} & \small CLEAR \cite{clear} & \small NeRT \cite{nert} & \small TriINR (ours) 
\end{tabular}
\caption{Qualitative comparison on the OTIS dataset~\cite{otis}, each restored from 20 input frames.}
\label{fig:qual_otis_odd}
\end{figure*}

\begin{figure*}[t]
\centering
\setlength{\tabcolsep}{1pt}
\renewcommand{\arraystretch}{0}
\begin{tabular}{@{}c@{\hspace{2pt}}cccccc@{}}
\adjustbox{valign=c,rotate=90}{\small \quad \quad \quad Pattern2} &
\includegraphics[width=0.15\linewidth]{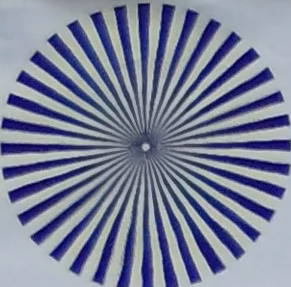}&
\includegraphics[width=0.15\linewidth]{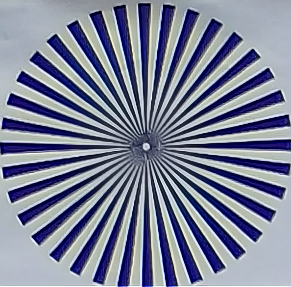}&
\includegraphics[width=0.15\linewidth]{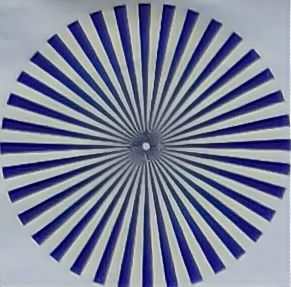}&
\includegraphics[width=0.15\linewidth]{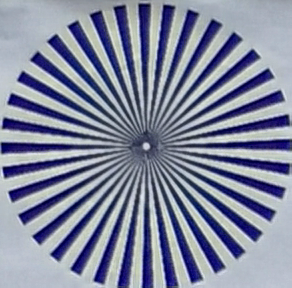}&
&
\includegraphics[width=0.15\linewidth]{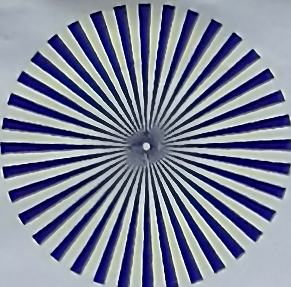}
\\[1pt]
\adjustbox{valign=c,rotate=90}{\small \quad \quad \quad Pattern4} &
\includegraphics[width=0.15\linewidth]{figsandtbs/qualitative_otis/Pattern4/input_turb.png}&
\includegraphics[width=0.15\linewidth]{figsandtbs/qualitative_otis/Pattern4/tmt.png}&
\includegraphics[width=0.15\linewidth]{figsandtbs/qualitative_otis/Pattern4/datum.png}&
\includegraphics[width=0.15\linewidth]{figsandtbs/qualitative_otis/Pattern4/clear.png}&
\includegraphics[width=0.15\linewidth]{figsandtbs/qualitative_otis/Pattern4/nertk11.png}&
\includegraphics[width=0.15\linewidth]{figsandtbs/qualitative_otis/Pattern4/pseudo_clean.png}
\\[1pt]
\adjustbox{valign=c,rotate=90}{\small \quad \quad \quad Pattern6} &
\includegraphics[width=0.15\linewidth]{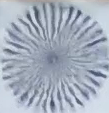}&
\includegraphics[width=0.15\linewidth]{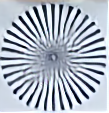}&
\includegraphics[width=0.15\linewidth]{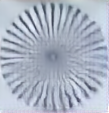}&
\includegraphics[width=0.15\linewidth]{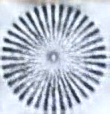}&
\includegraphics[width=0.15\linewidth]{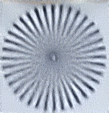}&
\includegraphics[width=0.15\linewidth]{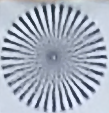}
\\[1pt]
\adjustbox{valign=c,rotate=90}{\small \quad \quad \quad Pattern8} &
\includegraphics[width=0.15\linewidth]{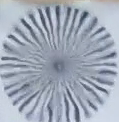}&
\includegraphics[width=0.15\linewidth]{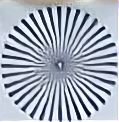}&
\includegraphics[width=0.15\linewidth]{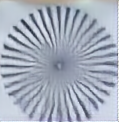}&
\includegraphics[width=0.15\linewidth]{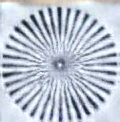}&
\includegraphics[width=0.15\linewidth]{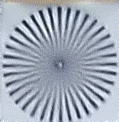}&
\includegraphics[width=0.15\linewidth]{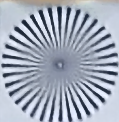}
\\[1pt]
\adjustbox{valign=c,rotate=90}{\small \quad \quad \quad Pattern10} &
\includegraphics[width=0.15\linewidth]{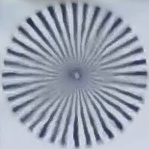}&
\includegraphics[width=0.15\linewidth]{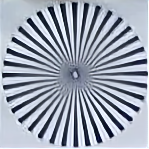}&
\includegraphics[width=0.15\linewidth]{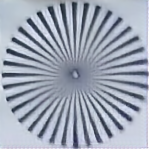}&
\includegraphics[width=0.15\linewidth]{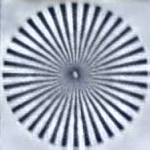}&
\includegraphics[width=0.15\linewidth]{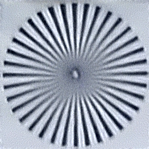}&
\includegraphics[width=0.15\linewidth]{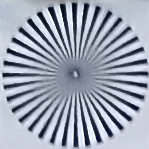}
\\[1pt]
\adjustbox{valign=c,rotate=90}{\small \quad \quad \quad Pattern12} &
\includegraphics[width=0.15\linewidth]{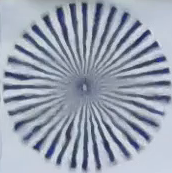}&
\includegraphics[width=0.15\linewidth]{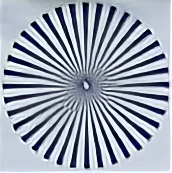}&
\includegraphics[width=0.15\linewidth]{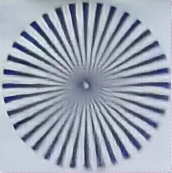}&
\includegraphics[width=0.15\linewidth]{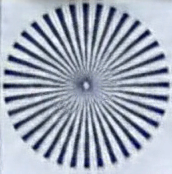}&
\includegraphics[width=0.15\linewidth]{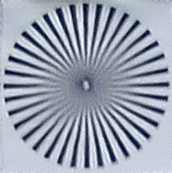}&
\includegraphics[width=0.15\linewidth]{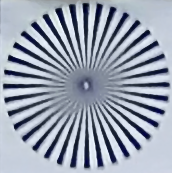}
\\[1pt]
\adjustbox{valign=c,rotate=90}{\small \quad \quad \quad Pattern14} &
\includegraphics[width=0.15\linewidth]{figsandtbs/qualitative_otis/Pattern14/input_turb.png}&
\includegraphics[width=0.15\linewidth]{figsandtbs/qualitative_otis/Pattern14/tmt.png}&
\includegraphics[width=0.15\linewidth]{figsandtbs/qualitative_otis/Pattern14/datum.png}&
\includegraphics[width=0.15\linewidth]{figsandtbs/qualitative_otis/Pattern14/clear.png}&
\includegraphics[width=0.15\linewidth]{figsandtbs/qualitative_otis/Pattern14/nertk11.png}&
\includegraphics[width=0.15\linewidth]{figsandtbs/qualitative_otis/Pattern14/pseudo_clean.png}
\\[1pt]
\adjustbox{valign=c,rotate=90}{\small \quad \quad \quad Pattern16} &
\includegraphics[width=0.15\linewidth]{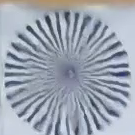}&
\includegraphics[width=0.15\linewidth]{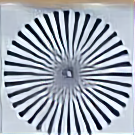}&
\includegraphics[width=0.15\linewidth]{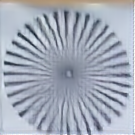}&
\includegraphics[width=0.15\linewidth]{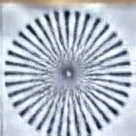}&
\includegraphics[width=0.15\linewidth]{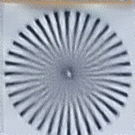}&
\includegraphics[width=0.15\linewidth]{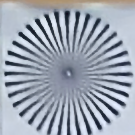}
\\[1pt]
&  \small Input sample & \small TMT \cite{tmt} & \small DATUM \cite{datum} & \small CLEAR \cite{clear} & \small NeRT \cite{nert} & \small TriINR (ours) 
\end{tabular}
\caption{Additional qualitative comparison on OTIS patterns~\cite{otis}, each restored from 20 input frames.}
\label{fig:qual_otis_even}
\end{figure*}

\end{document}